\documentclass[10pt,twocolumn,letterpaper]{article}

\usepackage[pagenumbers]{cvpr}

\usepackage{cuted}
\usepackage{colortbl}
\definecolor{vloopgreen}{HTML}{EAF5EB}

\usepackage{currfile}

\usepackage{caption}

\usepackage{xspace}
\newcommand{\method}{VLoop\xspace}
\newcommand{\alder}{ALDER\xspace}

\newcommand{\appref}[1]{Appendix~\ref{#1}}
\newcommand{\appsref}[2]{Appendices~\ref{#1} and~\ref{#2}}

\definecolor{cvprblue}{rgb}{0.21,0.49,0.74}
\usepackage[pagebackref,breaklinks,colorlinks,allcolors=cvprblue]{hyperref}

\definecolor{linkblue}{HTML}{0071BC}
\definecolor{burntorange}{HTML}{BF5700}

\hypersetup{colorlinks=true, citecolor=linkblue, linkcolor=linkblue, urlcolor=burntorange}

\def\confName{CVPR}
\def\confYear{2027}

\title{Thinking Deeper, Not Longer: Adaptive Looped Latent Reasoning for Video}

\author{
Joungbin An \qquad Mi Luo \qquad Kristen Grauman \\
The University of Texas at Austin
}

\begin{document}
\maketitle
\addtocontents{toc}{\protect\setcounter{tocdepth}{0}}
\begin{abstract}
Recent video reasoning models reason by thinking \emph{longer}, either generating textual rationales or appending latent tokens before they answer.
We introduce \method, a video-language model that instead thinks \emph{deeper}.
\method forwards the video through the model once and repeatedly loops the hidden states of the question and a few latent slots through its core layers, each time revisiting the frames it has already encoded.
Because only these few tokens are updated against the cached video, each additional loop increases reasoning depth while adding only 1.2\% to inference FLOPs.
We further introduce Adaptive Loop-DEpth Reinforcement (\alder), a reinforcement learning strategy that rewards correct answers at every reasoning depth and teaches the model when to think deeper.
Across six video benchmarks, \method improves on its base model by up to 7.2 points on temporal reasoning and by 3.9 points on average---more than four times the gain of text-CoT and latent-expansion methods with the same backbone and frame budget---while answering 5$\times$ faster than text-CoT.\footnote{Project webpage: \url{https://vision.cs.utexas.edu/projects/vloop}.}
\vspace{-4mm}
\end{abstract}

\begin{figure}[t]
\centering
\includegraphics[width=\linewidth]{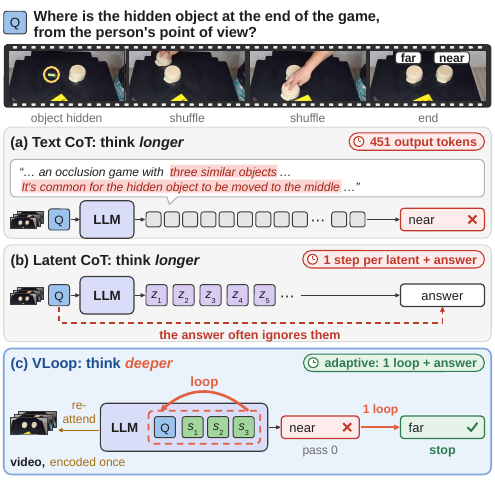}
\vspace{-7mm}
\caption{\textbf{Thinking deeper, not longer.}
(a) A reasoning-augmented VLM miscounts the objects and reasons from where the object is commonly hidden rather than from the video (red).
(b)~Sequential latent reasoning adds tokens (schematic), whose removal can leave answers largely unchanged~\cite{zhang2026ablate,luo2026anchorsim}.
(c)~\method encodes the video once and refines the question and latent slots.
In this example, one loop corrects the initial answer and the learned gate stops.
Clocks count reasoning steps: decoded tokens, latent tokens, or loops, respectively.
}
\label{fig:teaser}
\vspace{-3mm}
\end{figure}

\section{Introduction}
\label{sec:intro}

Watch a cup-and-ball trick.
Every swap is in plain view, so the difficulty is not seeing them: answering \emph{where the ball ends up} requires chaining the swaps, one after another, into the ball's final position.
A viewer does this not by rewatching the video or narrating each step, but instead by \emph{updating} a mental
model of the ball's location and stating only the conclusion.
As in action ordering and motion tracking, the relevant observations are available; the challenge is to reason over them in several steps.

Processing more frames or higher-resolution inputs can provide video-language models with richer visual evidence.
However, when the relevant evidence is already available, a complementary challenge is to reason over it in multiple steps.
Textual chain-of-thought (CoT) does so
by generating a reasoning trace token by token before answering~\cite{wei2022cot,huang2025visionr1,feng2025videor1,wang2025videorft,meng2025openo3}.
This sequential generation adds inference latency~\cite{hao2024coconut}, but does not guarantee a better answer: the resulting rationales can introduce unsupported visual details and drift from the observed evidence~\cite{luo2025drift,dugad2026vios}.

In light of these limitations, researchers are exploring
reasoning without verbalizing every intermediate step.
In particular, latent reasoning
carries out intermediate computation through continuous representations without decoding each step into words~\cite{hao2024coconut,yang2025mirage,li2025lvr,qin2025covt}.
This distinction between thought and its verbal expression is consistent with neuroscientific evidence that complex human reasoning need not depend on language~\cite{fedorenko2024language}.

Yet existing models fall short. Many latent reasoning methods think \emph{longer} through \emph{latent expansion}, extending the reasoning sequence with additional latent tokens, often generated one at a time~\cite{hao2024coconut,li2025lvr,ray2025mull,hu2026videolatent,xia2026dylar,liang2026torm}.
However, extending the reasoning sequence does not ensure that the generated states contribute to the answer: recent ablations identify settings in which perturbing latent content has little effect on answer accuracy or likelihood~\cite{zhang2026ablate,luo2026anchorsim}.

We introduce \method, a looped latent reasoner for video that thinks \emph{deeper} by revising an internal state, akin to updating a mental model
(\cref{fig:teaser}).
This state comprises the hidden states of the question tokens and a compact latent memory of fixed size.
After processing the video once, \method updates all question and memory positions in parallel within each loop through shared core decoder layers.
Their evolving representations guide renewed attention to the cached video evidence, producing the state used by the next loop.
Additional computation thus increases reasoning depth without extending the reasoning sequence or reprocessing the video.
\method learns these state updates directly from answer supervision, enabling training on standard video--question--answer triplets without annotated reasoning traces or prescribed latent targets.

While \method is designed to think deeper, it is also designed to know when its loops have thought deeply enough.
The value of additional depth
depends on the question and the current reasoning state.
Further refinement can correct an initial mistake, leave the answer unchanged, or turn a correct prediction into an error.
We therefore introduce Adaptive Loop-DEpth Reinforcement (\alder), which trains answer generation at every depth and learns whether to refine the current state or answer.
Its stopping objective rewards the correctness of the answer at the chosen exit, with a small per-loop penalty that favors answering earlier when refinement would not change the outcome.

With the same backbone and frame budget, \method outperforms representative text-CoT~\cite{feng2025videor1,luo2025drift} and latent-expansion~\cite{ray2025mull,hu2026videolatent,xia2026dylar} methods while answering faster (\cref{tab:main_results,fig:reasoning_latency}).
Under matched training, looping the question and latent slots is more accurate than the same-data SFT model or looping the latent slots alone, with the largest gains on temporal reasoning.
\alder further improves accuracy at every exit and learns when to think deeper, dynamically considering the question, video, and latent thought progress so far.

Our contributions are
(1) \method, to our knowledge the first looped latent reasoner for video, which increases computational depth by recurrently refining the hidden states of the question tokens and a fixed-size latent memory against the cached video; we find this choice of what to loop is what makes looping effective on video;
(2) \alder, which trains answer generation at every depth and learns when to think deeper and when to answer; and
(3) an analysis of when refinement helps or hurts, including matched-training comparisons and alternative looping schemes, alongside accuracy--latency comparisons with existing video reasoning methods.

\section{Related Work}
\label{sec:related_work}

\noindent\textbf{Video Reasoning.}
Vision-language models~\cite{chen2024internvl,li2024onevision,bai2025qwen25vl,openai2024gpt4o,tong2024cambrian} provide a foundation for reasoning about spatial relations~\cite{ray2024sat}, object transformations~\cite{yi2019clevrer,xue2024states}, and activity progression~\cite{tang2019coin,xue2025progress}.
Building on chain-of-thought (CoT) reasoning in language models~\cite{wei2022cot,wang2022selfconsistency}, multimodal approaches train models to generate textual rationales through supervision and reinforcement learning~\cite{huang2025visionr1,feng2025videor1,liu2026videoauto}.
However, these rationales can hallucinate visual details~\cite{zhang2025vital} or drift from the observed evidence~\cite{luo2025drift}.
To strengthen grounding, methods reward or distill evidence-consistent rationales~\cite{luo2025drift,dugad2026vios}, link reasoning steps to image regions~\cite{fan2025grit}, invoke visual tools~\cite{jiang2025vlmr3,zheng2025deepeyes,wang2025pixelreasoner,hong2025deepeyesv2}, or, for videos, interleave frames with text or retrieve additional clips and frames~\cite{zhang2026vticot,yang2025longvt,zhang2025vital}.
Even with relevant evidence, verbal descriptions can omit spatial and temporal detail~\cite{yang2025mirage,zhang2026vticot}.

Visual intermediates offer a complementary representation.
Alongside unified models for visual understanding and generation~\cite{wu2024vilau,deng2025bagel,liu2025javisgpt}, methods reason through intermediate images and image--text traces~\cite{ou2025d2r,li2025mvot,gu2025thinkmorph,li2025zebracot,xu2025visualplanning} or generated videos~\cite{wiedemer2025videoreasoners,tong2025thinkingvideo,wang2026vbvr,wang2026demystifying}.
Latent video reasoners pursue intermediate visual computation without rendering these states~\cite{hu2026videolatent,liang2026torm,xia2026dylar}.

\noindent\textbf{Continuous Latent Reasoning.}
Text-only language models can reason without verbalizing every intermediate step, using pause tokens~\cite{goyal2023pause}, continuous-state feedback~\cite{hao2024coconut}, or CoT distilled into latent states~\cite{shen2025codi}.
For vision, continuous representations can retain details omitted by verbal descriptions~\cite{yang2025mirage,li2025lvr,zhang2025sketchpad} and are trained with intermediate-image, regional, or expert-feature targets~\cite{yang2025mirage,li2025lvr,qin2025covt}, reconstruction or distillation~\cite{tong2025skila,dong2025ilvr}, or reinforcement learning~\cite{wang2025monet}.
Discrete visual and modality-agnostic tokens provide alternative representations for nonverbal reasoning~\cite{bigverdi2025perception,ray2025mull}.

Many methods perform multistep reasoning through \emph{latent expansion}, extending the reasoning sequence with latent states.
A common implementation appends these states between boundary tokens before answering~\cite{hao2024coconut,yang2025mirage,li2025lvr,shen2025codi}.
Video methods guide them with generated thought-video targets~\cite{liang2026torm}, compressed frame--text traces~\cite{zhang2026vticot}, key-frame features~\cite{zhang2026latentvc}, or rationale distillation with an optional reasoning block~\cite{xia2026dylar}, or learn them using a dedicated module that injects video and question features~\cite{hu2026videolatent}.

Latent expansion, however, does not guarantee that the added states contribute to the answer.
Recent ablations show that accuracy gains can persist after latent replacement at a fixed token budget~\cite{zhang2026ablate}, and contrastive distillation aims to strengthen answer dependence on visual latents~\cite{luo2026anchorsim}.
We study recurrent refinement of a mental model as a complementary form of multistep latent reasoning.

\noindent\textbf{Looped Transformers and Recurrent Reasoning.}
Recurrent architectures offer an alternative to latent expansion: they increase computational depth by repeatedly applying shared layers to refine existing states.
This approach has been explored in natural language processing and algorithmic reasoning~\cite{dehghani2018universal,geiping2025huginn,saunshi2025looped}.
CoT-supervised variants align iterations with textual reasoning steps~\cite{yu2026relay} or refine fixed latent blocks against a cached question prefix~\cite{fan2026lotus}.
Related work explores compact recursive networks for structured puzzles~\cite{wang2025hrm,jolicoeurmartineau2025trm}.
Because excessive recurrence can degrade predictions~\cite{bansal2022overthinking,kohli2026loopthink}, some of these models adapt computation to each input through learned halting~\cite{graves2016act,banino2021pondernet,zhu2025ouro} or recursive routing~\cite{bae2025mor}.
Image-based work also revisits fixed visual evidence, then discards the original multimodal cache so that the recurrent state is the sole visual input to answering~\cite{park2026cvrr}.
Less work considers looping in vision-language models, which recur over every image and text token~\cite{qian2026loopvl,xu2026recursivevlm}; latent-block recurrence in language models instead refines only appended latents against a fixed prefix~\cite{fan2026lotus}.

To our knowledge, \method is the first looped transformer for video.  We find that neither existing scheme suits video: looping every token costs 24 times more per loop for the same accuracy, and looping latent slots alone does not help (\cref{sec:depth_ablation}).
Inspired by mental imagery, \method instead updates only the hidden states of the question tokens and a fixed-size latent memory through renewed attention to encoded frames.
Every loop re-attends to the video's key--value cache, computed once and reused, and the question itself recurs. In contrast,
the closest image-based reasoner discards this cache so that it answers from the recurrent state alone~\cite{park2026cvrr}.
Because only these few states are recomputed, each loop adds only 1.2\% inference FLOPs.
The model learns refinement directly from video--question--answer supervision through the backbone's existing attention, without CoT supervision, prescribed intermediate reasoning targets, or a dedicated latent-injection module.
Unlike learned halting~\cite{graves2016act,banino2021pondernet,zhu2025ouro}, which is trained through the task loss, \alder rewards the correctness of the answer at the chosen exit, so the policy also learns to stop before a loop overturns a correct answer.

\section{How Can Video Models Think Deeper?}
\label{sec:looped_reasoning}

We examine how existing methods allocate computation before answering, then introduce recurrent refinement as a way to think deeper over encoded video evidence.

\begin{figure}[!t]
\centering
\includegraphics[width=\linewidth]{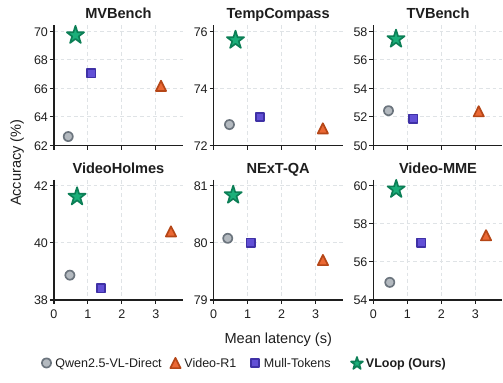}
\vspace{-9mm}
\caption{\textbf{Accuracy versus latency.}
Each panel plots benchmark accuracy against mean latency per question (16 frames, batch size one on an H100 GPU); upper left is better.
Qwen2.5-VL is the base model in each case.
}
\label{fig:reasoning_latency}
\vspace{-3mm}
\end{figure}

\begin{figure*}[!t]
\centering
\includegraphics[width=\linewidth]{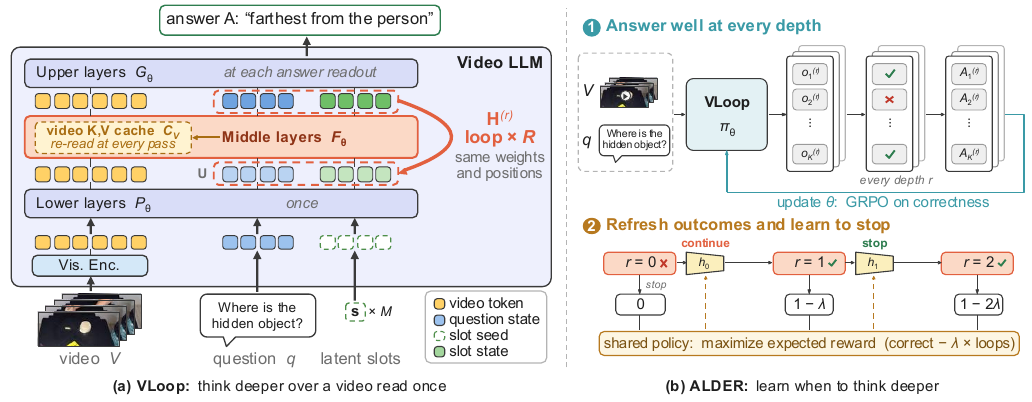}
\vspace{-8mm}
\caption{\textbf{\method: recurrent reasoning with adaptive depth.}
(a)~Video tokens are processed once; only the question and latent slots loop through $F_\theta$, reusing the video cache $\mathcal{C}_V$ at fixed sequence length.
(b)~\alder rewards correct answers at every depth and uses fresh outcomes to train a shared stopping policy to maximize expected answer correctness minus loop cost.
}
\label{fig:method}
\vspace{-5mm}
\end{figure*}

\subsection{When Does Additional Reasoning Help?}
\label{sec:reasoning_utility}

Textual CoT methods allocate additional computation by generating a textual rationale before answering~\cite{feng2025videor1}.
Latent expansion methods use additional nonverbal states, generated sequentially~\cite{hu2026videolatent,xia2026dylar} or introduced as a fixed block processed jointly~\cite{ray2025mull}.
Both approaches devote intermediate reasoning to additional sequence positions.

\Cref{fig:reasoning_latency} compares direct answering with a representative method from each family: Video-R1~\cite{feng2025videor1} for text CoT and Mull-Tokens~\cite{ray2025mull} for latent expansion, both built on the same backbone~\cite{bai2025qwen25vl} and given identical sampled frames.
Their benefit is mixed: relative to direct answering, each gains on some benchmarks but loses accuracy on others.
Their cost, in contrast, is consistently higher: mean latency is about 6.9--7.8 times that of direct answering for Video-R1 and 2.6--3.0 times for Mull-Tokens.

One likely reason is that questions differ in the reasoning they need, and benchmarks mix them in different proportions.
Some can be answered at a glance, such as naming an object or an action; there, additional reasoning adds cost and can drift from the evidence~\cite{luo2025drift}.
Others require relating observations across time or space, such as tracking a hidden ball through a series of swaps or judging the direction of a motion.
For these, the needed evidence is already encoded; what remains is to connect it, through relations that words describe only coarsely~\cite{yang2025mirage,zhang2026vticot}.

This suggests reasoning in continuous space over the representations the model has already formed, much as people maintain and manipulate visual information in working memory~\cite{pearson2019workingmemory}.
Adding sequence positions is a costly way to do so: Video-R1 decodes hundreds of tokens one at a time, and latent states added to the sequence need not be used by the answer~\cite{zhang2026ablate,luo2026anchorsim}.
We therefore refine an existing reasoning state rather than extend the sequence, and, since certain questions
need little such refinement, spend it only where it helps (\cref{sec:depth_diagnostic}).

Concretely, \method updates the hidden states of the question tokens together with a fixed-size latent memory: these question states can adapt their attention to the observed video, while the memory carries information across passes and integrates it with this evolving question context.
It implements this refinement by repeatedly applying a shared middle-layer core to only this compact segment (\cref{fig:method}(a)).
The much longer video prefix is processed once, and its key--value cache supplies the visual evidence for every pass.

\subsection{Looped Latent Reasoning for Video}
\label{sec:latent_state}
\label{sec:recurrent_refinement}

\noindent\textbf{Video cache and latent state.}
We use a pretrained video-language model with a visual encoder, projector, and $L$-layer causal decoder of hidden dimension $d$.
Given a video $V$ and question $q$, the visual encoder and projector encode the sampled frames once.
A standard decoder forward pass over the video prefix computes the per-layer self-attention keys and values, which we retain as $\mathcal{C}_V$ throughout refinement and answering.

Let $\mathbf{E}_q\in\mathbb{R}^{n_q\times d}$ be the embeddings of the question that follows the video tokens.
We append a latent-start embedding $\mathbf{e}_{\mathrm{start}}$ and $M$ latent slots:
\vspace{-2mm}
\begin{equation}
    \mathbf{X} = [\mathbf{E}_q;\,\mathbf{e}_{\mathrm{start}};\,\mathbf{S}],
    \qquad \mathbf{S} = [\mathbf{s};\,\ldots;\,\mathbf{s}],
    \label{eq:latent_state}
    \vspace{-2mm}
\end{equation}
where semicolons concatenate sequence positions and $\mathbf{S}$ repeats a shared trainable seed $\mathbf{s}\in\mathbb{R}^d$ at $M$ positions.
Distinct positions allow the identically initialized slots to develop different representations through attention to the question and video cache.

\noindent\textbf{Recurrent refinement.}
Layer-wise probing of MLLMs associates early layers with visual grounding and lexical integration, middle layers with semantic reasoning, and late layers with answer decoding~\cite{yu2025multimodallayers}.
Motivated by this organization, we partition the decoder into lower, recurrent middle, and upper blocks $P_\theta$, $F_\theta$, and $G_\theta$ (\cref{fig:method}(a)).
For $R$ additional core passes, the computation is
\vspace{-2mm}

\begin{equation}
    \begin{aligned}
        \mathbf{U} &= P_\theta(\mathbf{X};\mathcal{C}_V),\\
        \mathbf{H}^{(0)} &= F_\theta(\mathbf{U};\mathcal{C}_V),\\
        \mathbf{H}^{(r)} &= F_\theta(\mathbf{H}^{(r-1)};\mathcal{C}_V),\\
        \mathbf{O}^{(R)} &= G_\theta(\mathbf{H}^{(R)};\mathcal{C}_V).
    \end{aligned}
    \label{eq:looped_refinement}
    \vspace{-1mm}
\end{equation}
Each block attends to the video cache at its corresponding layers.
The initial core pass forms $\mathbf{H}^{(0)}$, followed by $R$ additional passes indexed by $r=1,\ldots,R$.
Thus, $R=0$ already processes the question and latent memory once; it denotes reasoning without additional loops.

Feeding $\mathbf{H}^{(r-1)}$ back into the core produces new attention queries from the current state, enabling later passes to reweight visual evidence and combine observations across frames.
Each additional loop processes the same $n_q+1+M$ positions through $F_\theta$, jointly within the pass; successive passes remain sequential.
These positions still attend to the cached video, but video-token projections and feed-forward computations are not repeated.
Earlier-pass keys and values are not appended to the attention context, so its length stays fixed as depth increases.

\noindent\textbf{Answer generation and supervised initialization.}
After $r$ additional passes (exit $r$), we apply $G_\theta$, append a latent-end boundary, and generate the answer $y=(y_1,\ldots,y_T)$ autoregressively; we write $p_\theta(y\mid V,q;r)$ for its distribution.
Each answer token traverses the full decoder once, attending to the video cache and to the question-and-slot keys and values, which come from the final pass in $F_\theta$ and from the single pass in $P_\theta$ and $G_\theta$.
We initialize the model by supervised fine-tuning (SFT) on video--question--answer triplets $(V,q,y)\sim\mathcal{D}$ at a fixed depth $R_{\mathrm{SFT}}$:
\vspace{-1mm}
\begin{equation}
    \mathcal{L}_{\mathrm{SFT}}
    =-\mathbb{E}_{\mathcal{D}}\!\left[
        \frac{1}{T}\sum_{t=1}^{T}
        \log p_\theta(y_t\mid y_{<t},V,q;R_{\mathrm{SFT}})
    \right].
    \label{eq:answer_sft}
    \vspace{-1mm}
\end{equation}
Only answer tokens at the final exit have targets; their loss propagates through all recurrent passes, supervising the state updates through their effect on the answer without prescribed intermediate targets, and the model does not yet learn when to stop.
\Cref{sec:adaptive_depth} examines when further refinement warrants its cost, then introduces \alder to reward correct answers at every exit and
learn whether to refine further or answer (\cref{fig:method}(b)).

\section{Adaptive Loop-Depth Reinforcement}
\label{sec:adaptive_depth}
\vspace{-1mm}
\Cref{sec:reasoning_utility} suggested that questions differ in how much reasoning they need. In \method, each loop is one step of reasoning, so questions differ in how many loops they need: an initial pass already answers some, while others need further refinement.
Yet another loop can also overturn a correct answer, much as a test taker who second-guesses can talk themselves out of a right one.
The challenge is therefore to recognize, for each question and current reasoning state, when another loop is worthwhile.
\vspace{-1mm}

\subsection{How Does Answer Quality Change with Depth?}
\label{sec:depth_diagnostic}
\begin{figure}[t]
\centering
\includegraphics[width=\linewidth]{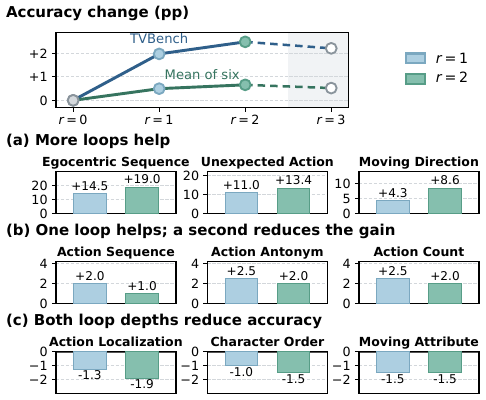}
\vspace{-7mm}
\caption{\textbf{Additional depth helps on average, but video tasks favor different depths.}
Accuracy change (pp) from $r=0$ of the SFT model at fixed depths, before \alder.
Top: unweighted mean over six benchmarks, and TVBench, at $r=1,2$ and $r=3$ (shaded). (a)--(c): subcategories of temporal reasoning benchmarks~\cite{cores2024tvbench,li2024mvbench}:
(a) gains at both loops; (b) first-loop gains reduced by the second; (c) losses at both depths.
}
\label{fig:sft_depth_combined}
\vspace{-5mm}
\end{figure}

\vspace{-1mm}
We evaluate the SFT initializer at $r\in\{0,1,2\}$ on six benchmarks.
Since $r=0$ already forms a question-conditioned latent state, much like latent expansion with a fixed block of latent tokens~\cite{ray2025mull}, these comparisons measure the value of further refinement over the same evidence.
\Cref{fig:sft_depth_combined} (top) shows that mean accuracy rises with each loop with diminishing returns, and rises more on benchmarks whose questions are built to require temporal information~\cite{cores2024tvbench} than on those that mix in general questions.
Yet this average does not imply that every question benefits.
\Cref{fig:sft_depth_combined}(a)--(c) examines this variation through three patterns in subcategories of two temporal reasoning benchmarks~\cite{cores2024tvbench,li2024mvbench}, chosen within each pattern by a fixed ranking of their accuracy change; \appref{sec:full_depth_analysis} provides results for all categories. In particular, the largest gains are on categories such as egocentric sequence, unexpected action, and moving direction (\cref{fig:sft_depth_combined}a).

\noindent\textbf{Refinement can help connect evidence across time.}
Row (a) shows successive gains on tasks such as ordering events and tracking motion direction, which require combining observations across frames.

\noindent\textbf{More thinking does not always improve the answer.}
In row (b), a first loop helps with ordering, counting, or contrasting actions, but a second reduces that gain, as if refinement moves past a useful state.
The losses in row (c) involve questions decided by a precise detail rather than by relating events: when an action occurs, the order of written characters, or the attribute of a moving object.
Refinement reuses the evidence already encoded, so it may not recover a detail the first pass missed, and further updates may drift from it.
Temporal reasoning therefore does not imply that the deepest exit is best.

\noindent\textbf{More thinking should be decided per question.}
The usefulness of further thinking depends on the question, not just the task.
Across the six benchmarks, two loops correct many answers that are wrong at $r=0$ but overturn many that are right, so no fixed depth suits every question.
Adaptive depth therefore needs an answer at every exit and a decision, made from each question’s current state, about whether to think further.

\subsection{Training \alder}
\label{sec:ald_training}
\vspace{-1mm}
\alder provides both answers at every exit and a stopping policy that sets the depth for each question, starting from the SFT model.
On the same training questions (\cref{sec:setup}), it interleaves two objectives (\cref{fig:method}(b)): answer training updates the backbone and latent seed so that every exit answers well (\cref{sec:all_depth_answers}), and stopping training updates only a small head that decides when to stop (\cref{sec:learned_stopping}).
\vspace{-1mm}

\subsubsection{Learning to Answer at Every Depth}
\label{sec:all_depth_answers}
\vspace{-1mm}
Questions that need several loops will not be answered correctly at an early exit, but stopping early pays off on those that can, provided that exit has learned to answer.
We therefore extend answer training from the final exit used in SFT to every exit, for every question and independently of the stopping policy: early exits learn to answer directly, while deeper exits keep receiving supervision even when the policy would stop before reaching them.

Let $R\geq1$ be the maximum additional depth and $x=(V,q)$ an input.
At each exit $r\in\{0,\ldots,R\}$, we sample $K\geq2$ answers from the behavior policy, with binary correctness indicators $c_i^{(r)}$.
Each answer is compared with the mean of the other samples at the same depth, a leave-one-out baseline~\cite{ahmadian2024rloo} without standard-deviation normalization:
\vspace{-1mm}
\begin{equation}
    A_i^{(r)}=c_i^{(r)}-\frac{1}{K-1}\sum_{j\ne i}c_j^{(r)}.
    \label{eq:depth_advantage}
    \vspace{-1mm}
\end{equation}
If all samples at an exit are wrong, their advantages vanish, so an early exit is not pushed toward answers it cannot yet reach.
We optimize these advantages with a clipped GRPO objective~\cite{shao2024deepseekmath}, averaging over sampled answers and weighting all depths equally.
Two regularizers encourage retention of the initializer's early-answer behavior and invariance to permutations of the answer choices:
\vspace{-1.5mm}
\begin{equation}
    \mathcal{L}_{\mathrm{ans}}=
    \mathcal{L}_{\mathrm{GRPO}}+\beta_{\mathrm{ref}}\mathcal{L}_{\mathrm{ref}}
    +\beta_{\mathrm{perm}}\mathcal{L}_{\mathrm{perm}}.
    \label{eq:ald_answer_loss}
    \vspace{-1.5mm}
\end{equation}
Here, $\mathcal{L}_{\mathrm{ref}}$ is KL from the frozen SFT answer distribution to the current model at $r=0$, and $\mathcal{L}_{\mathrm{perm}}$ sums, across exits, the KL between answer distributions under the original and a permuted order of the answer choices.

\begin{table*}[!t]
\centering
\caption{\textbf{Main results.} \method is the most accurate on every benchmark, by the widest margins on temporal reasoning.
Accuracy (\%) of 7B models at 16 frames (32 frames in \appref{sec:app_32frames}); bold marks the best per column. For reasoning baselines with public code and checkpoints, benchmarks not reported in prior work are evaluated by us with that code and checkpoint; all other reasoning-baseline scores are published. --: not reported in prior work, and no released checkpoint to evaluate. $^\dagger$: SFT with the same training data as \method. Green rows are ours: SFT only uses $r=2$; the final model (Ours) adds \alder and uses adaptive depth. Green: change from base Qwen2.5-VL with direct answering at the same frame budget.
}
\vspace{-3mm}
\label{tab:main_results}
\providecolor{vloopgain}{HTML}{36745F}
\footnotesize
\setlength{\tabcolsep}{4pt}
\renewcommand{\arraystretch}{0.92}
\begin{tabular}{llrrrrrrr}
\toprule
 & & \multicolumn{3}{c}{Temporal reasoning} & \multicolumn{2}{c}{Complex \& causal reasoning} & \multicolumn{1}{c}{General} & \\
\cmidrule(lr){3-5}\cmidrule(lr){6-7}\cmidrule(lr){8-8}
Method & Reasoning & \multicolumn{1}{c}{MVBench} & \multicolumn{1}{c}{TempCompass} & \multicolumn{1}{c}{TVBench} & \multicolumn{1}{c}{VideoHolmes} & \multicolumn{1}{c}{NExT-QA} & \multicolumn{1}{c}{Video-MME} & \multicolumn{1}{c}{Avg.} \\
\midrule
Qwen2.5-VL~\cite{bai2025qwen25vl} & Direct & 62.6\phantom{\,{\tiny(+7.2)}} & 72.7\phantom{\,{\tiny(+7.2)}} & 52.4\phantom{\,{\tiny(+7.2)}} & 38.9\phantom{\,{\tiny(+7.2)}} & 80.1\phantom{\,{\tiny(+7.2)}} & 54.9\phantom{\,{\tiny(+7.2)}} & 60.3\phantom{\,{\tiny(+7.2)}} \\
Video-R1~\cite{feng2025videor1} & Text CoT & 62.7\phantom{\,{\tiny(+7.2)}} & 72.6\phantom{\,{\tiny(+7.2)}} & 52.4\phantom{\,{\tiny(+7.2)}} & 40.4\phantom{\,{\tiny(+7.2)}} & 79.7\phantom{\,{\tiny(+7.2)}} & 57.4\phantom{\,{\tiny(+7.2)}} & 60.9\phantom{\,{\tiny(+7.2)}} \\
VideoRFT~\cite{wang2025videorft} & Text CoT & 62.7\phantom{\,{\tiny(+7.2)}} & 70.8\phantom{\,{\tiny(+7.2)}} & 52.8\phantom{\,{\tiny(+7.2)}} & 39.8\phantom{\,{\tiny(+7.2)}} & 75.9\phantom{\,{\tiny(+7.2)}} & 57.5\phantom{\,{\tiny(+7.2)}} & 59.9\phantom{\,{\tiny(+7.2)}} \\
Video-VER~\cite{luo2025drift} & Text CoT & 63.2\phantom{\,{\tiny(+7.2)}} & 72.8\phantom{\,{\tiny(+7.2)}} & 51.0\phantom{\,{\tiny(+7.2)}} & --\phantom{\,{\tiny(+7.2)}} & --\phantom{\,{\tiny(+7.2)}} & 56.0\phantom{\,{\tiny(+7.2)}} & --\phantom{\,{\tiny(+7.2)}} \\
Open-o3-Video~\cite{meng2025openo3} & Text CoT & 63.7\phantom{\,{\tiny(+7.2)}} & 70.7\phantom{\,{\tiny(+7.2)}} & 52.8\phantom{\,{\tiny(+7.2)}} & 40.3\phantom{\,{\tiny(+7.2)}} & 79.2\phantom{\,{\tiny(+7.2)}} & 59.3\phantom{\,{\tiny(+7.2)}} & 61.0\phantom{\,{\tiny(+7.2)}} \\
LVR~\cite{li2025lvr} & Latent Expansion & 64.6\phantom{\,{\tiny(+7.2)}} & 72.0\phantom{\,{\tiny(+7.2)}} & 49.5\phantom{\,{\tiny(+7.2)}} & 33.4\phantom{\,{\tiny(+7.2)}} & 80.5\phantom{\,{\tiny(+7.2)}} & 59.1\phantom{\,{\tiny(+7.2)}} & 59.8\phantom{\,{\tiny(+7.2)}} \\
Mull-Tokens~\cite{ray2025mull} & Latent Expansion & 67.0\phantom{\,{\tiny(+7.2)}} & 73.0\phantom{\,{\tiny(+7.2)}} & 51.9\phantom{\,{\tiny(+7.2)}} & 38.4\phantom{\,{\tiny(+7.2)}} & 80.0\phantom{\,{\tiny(+7.2)}} & 57.0\phantom{\,{\tiny(+7.2)}} & 61.2\phantom{\,{\tiny(+7.2)}} \\
VideoLatent~\cite{hu2026videolatent} & Latent Expansion & 68.3\phantom{\,{\tiny(+7.2)}} & 73.7\phantom{\,{\tiny(+7.2)}} & --\phantom{\,{\tiny(+7.2)}} & 40.5\phantom{\,{\tiny(+7.2)}} & 80.7\phantom{\,{\tiny(+7.2)}} & 59.3\phantom{\,{\tiny(+7.2)}} & --\phantom{\,{\tiny(+7.2)}} \\
DyLaR~\cite{xia2026dylar} & Latent Expansion & 66.3\phantom{\,{\tiny(+7.2)}} & 72.4\phantom{\,{\tiny(+7.2)}} & --\phantom{\,{\tiny(+7.2)}} & 41.0\phantom{\,{\tiny(+7.2)}} & --\phantom{\,{\tiny(+7.2)}} & 59.8\phantom{\,{\tiny(+7.2)}} & --\phantom{\,{\tiny(+7.2)}} \\
\cmidrule(lr){1-9}
Qwen2.5-VL-SFT$^\dagger$ & Direct & 65.6\phantom{\,{\tiny(+7.2)}} & 73.7\phantom{\,{\tiny(+7.2)}} & 51.8\phantom{\,{\tiny(+7.2)}} & 40.0\phantom{\,{\tiny(+7.2)}} & 78.6\phantom{\,{\tiny(+7.2)}} & 57.6\phantom{\,{\tiny(+7.2)}} & 61.2\phantom{\,{\tiny(+7.2)}} \\
\rowcolor{vloopgreen}
\textbf{\method (SFT only)} & Latent Loop & 68.8\phantom{\,{\tiny(+7.2)}} & 74.5\phantom{\,{\tiny(+7.2)}} & 57.3\phantom{\,{\tiny(+7.2)}} & 40.6\phantom{\,{\tiny(+7.2)}} & 80.3\phantom{\,{\tiny(+7.2)}} & 59.5\phantom{\,{\tiny(+7.2)}} & 63.5\phantom{\,{\tiny(+7.2)}} \\
\rowcolor{vloopgreen}
\textbf{\method (Ours)} & Latent Loop & \textbf{69.8}\rlap{\,{\tiny\color{vloopgain}(+7.2)}}\phantom{\,{\tiny(+7.2)}} & \textbf{75.7}\rlap{\,{\tiny\color{vloopgain}(+3.0)}}\phantom{\,{\tiny(+7.2)}} & \textbf{57.5}\rlap{\,{\tiny\color{vloopgain}(+5.1)}}\phantom{\,{\tiny(+7.2)}} & \textbf{41.6}\rlap{\,{\tiny\color{vloopgain}(+2.7)}}\phantom{\,{\tiny(+7.2)}} & \textbf{80.9}\rlap{\,{\tiny\color{vloopgain}(+0.8)}}\phantom{\,{\tiny(+7.2)}} & \textbf{59.9}\rlap{\,{\tiny\color{vloopgain}(+5.0)}}\phantom{\,{\tiny(+7.2)}} & \textbf{64.2}\rlap{\,{\tiny\color{vloopgain}(+3.9)}}\phantom{\,{\tiny(+7.2)}} \\
\bottomrule
\end{tabular}
\vspace{-5mm}
\end{table*}

\subsubsection{Learning When to Stop}
\label{sec:learned_stopping}

The stopping decision must be made before the next loop is computed.
A small head shared across nonterminal depths predicts a continuation probability $h_r=\sigma(f_\phi(\mathbf{z}_r,r))$.
The detached features $\mathbf{z}_r$ are computed from the probabilities of the offered answers: three confidence measures (maximum probability, entropy, and top-two margin) and five measures of change from the preceding pass, which are zero at $r=0$ (\appref{sec:ald_stopping_details}).
The depth input distinguishes how much refinement has already occurred.

During training, greedy answers at every depth provide correctness indicators $\hat c_r$.
We reward the answer at the selected exit and penalize additional loops:
\vspace{-2mm}
\begin{equation}
    U_r=\hat c_r-\lambda r,
    \label{eq:stopping_utility}
    \vspace{-2mm}
\end{equation}
where a small $\lambda>0$ breaks ties toward earlier answers: equal correctness favors stopping, while any correction outweighs the penalty.
Evaluating all depths in training gives the expected return over every stopping choice:
\vspace{-2mm}
\begin{equation}
    J_\phi=\sum_{r=0}^{R}p_\phi(r)\,U_r,\;\;
    p_\phi(r)=(1-h_r)\prod_{j=0}^{r-1}h_j.
    \label{eq:stopping_return}
    \vspace{-2mm}
\end{equation}
The probability $p_\phi(r)$ accounts for continuing through earlier boundaries and stopping at $r$; we set $h_R=0$ and take the empty product to be one.
This halting distribution has the same form as in PonderNet~\cite{banino2021pondernet}, which weights the task loss of every step by it and so trains predictions and halting jointly.
We instead train every exit independently of stopping (\cref{sec:all_depth_answers}) and use the distribution only to optimize a small head for answer correctness minus
loop cost.
Rewarding the eventual answer allows the policy to value corrections requiring multiple loops, even if the next pass remains wrong.
We differentiate this expectation directly rather than sample stopping actions.
The head is updated online during RL on the current model's answers (\appref{sec:ald_stopping_details}).
\vspace{-1mm}
\subsection{Adaptive Inference}
\label{sec:adaptive_inference}
\vspace{-2mm}
Inference begins at $r=0$ and visits only the depth requested by the stopping policy.
At each reached exit, we apply $G_\theta$; at nonterminal exits, we also score the offered answers to compute the stopping features.
For a fixed threshold $\tau\in(0,1)$, we pass the current core state $\mathbf{H}^{(r)}$ through $F_\theta$ again only if $h_r>\tau$.
Otherwise, or upon reaching $r=R$, we answer from that exit; ties stop.
Training maximizes the expected return of a stochastic stopping policy (\cref{eq:stopping_return});
this deterministic rule takes its more likely action at each exit.
\alder thus lets \method find the best answer for each question, learning from the outcome at every exit when to think deeper and when not to.
\vspace{-2mm}

\section{Experiments}
\label{sec:experiments}
\vspace{-1.5mm}
\noindent\textbf{Implementation details.}
\label{sec:setup}
\method builds on Qwen2.5-VL-7B-Instruct~\cite{bai2025qwen25vl}, with decoder layers 10--20 as the recurrent core~\cite{yu2025multimodallayers} and $M=8$ latent slots.
For SFT, we use the multiple-choice questions and answers of the Video-R1-CoT corpus~\cite{feng2025videor1}, without their textual rationales.
SFT trains the language model, projector, and latent seed with the visual encoder frozen, using 16 frames, $R_{\mathrm{SFT}}=2$, AdamW with learning rate $10^{-6}$, and global batch size 66 for two epochs.
Following Video-R1~\cite{feng2025videor1}, each frame has at most $128\times28\times28$ pixels during training and $256\times28\times28$ pixels for evaluation.
For \alder, we profile questions from the training splits of video question-answering datasets~\cite{wu2021star,yi2019clevrer,patraucean2023perception,chen2025seedbenchr1,fu2026videokr,du2024reversed} with a \method SFT model and keep about 15K on which it is neither confidently right nor confidently wrong; none of their videos appear in the evaluation benchmarks.
\alder starts from the SFT model and trains for one epoch (635 training steps) with $R=2$, $K=8$ sampled answers per exit, learning rate $10^{-6}$, and global batch size 24.
Every 16 steps, the stopping head is updated on all-depth outcomes recomputed for the last 64 batches, with $\lambda=10^{-3}$; inference uses the threshold $\tau=0.5$.
Both stages train on six GH200, and latency is measured on a single H100 with batch size one.
Architecture, data, and training details are in \appsref{sec:looped_implementation}{sec:ald_implementation}.

\noindent\textbf{Benchmarks.}
Following prior video reasoning work~\cite{feng2025videor1,hu2026videolatent}, we evaluate on six video question answering benchmarks covering temporal reasoning (MVBench~\cite{li2024mvbench}, TempCompass~\cite{liu2024tempcompass}, TVBench~\cite{cores2024tvbench}), complex and causal reasoning (VideoHolmes~\cite{cheng2025videoholmes}, NExT-QA~\cite{xiao2021nextqa}), and general video understanding (Video-MME~\cite{fu2025videomme}).
We report accuracy and its unweighted mean over the six benchmarks.

\noindent\textbf{Baselines.}
We compare with two representative groups of 7B video reasoning models.
Text CoT methods generate a textual rationale before answering: Video-R1~\cite{feng2025videor1}, VideoRFT~\cite{wang2025videorft}, Video-VER~\cite{luo2025drift}, Open-o3-Video~\cite{meng2025openo3}, and VIOS~\cite{dugad2026vios}.
Latent expansion methods add latent tokens to the reasoning sequence: LVR~\cite{li2025lvr}, Mull-Tokens~\cite{ray2025mull}, VideoLatent~\cite{hu2026videolatent}, DyLaR~\cite{xia2026dylar}, STORM~\cite{liang2026torm}, and VTI-CoT~\cite{zhang2026vticot}.
We also report Qwen2.5-VL with direct answering and Qwen2.5-VL-SFT, a model trained on the same SFT data as \method.
For reasoning baselines, we use published scores where available; for those that publicly release code and checkpoints, we evaluate them under identical settings to fill missing entries.
VIOS, STORM, and VTI-CoT report only 32-frame results and release no checkpoints, so we compare with them at 32 frames (\appref{sec:app_32frames}).

\subsection{Main Results}
\label{sec:reasoning_comparison}
\vspace{-1mm}
\noindent\textbf{Comparison with prior methods.}
At 16 frames, \method achieves the best accuracy on all six benchmarks and the highest mean, 64.2\%, a gain of 3.9 points over direct Qwen2.5-VL that is \textbf{\emph{more than four times the largest gain}} among reasoning baselines with all six scores (\cref{tab:main_results}).
The margins are largest on the temporal reasoning benchmarks: on TVBench, \method is 4.7 points above the best baseline, versus 0.2 and 0.1 on NExT-QA and Video-MME.
At 32 frames, \method again has the highest mean (\appref{sec:app_32frames}).

\begin{figure}[t]
  \centering
  \includegraphics[width=\linewidth]{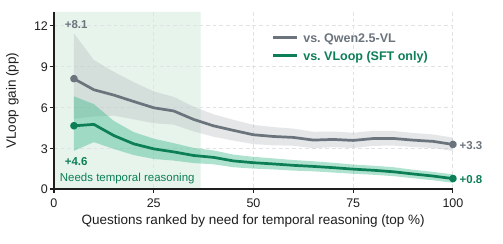}
  \vspace{-9.5mm}
  \caption{\textbf{\method's gains are largest on questions that need temporal reasoning.}
  Gain over Qwen2.5-VL and over \method (SFT only) on the $k\%$ of questions that most need temporal reasoning, pooled over the six benchmarks; shading marks the 37\% labeled as needing it.
  Bands: 95\% bootstrap intervals over videos.
  }
  \label{fig:reasoning_need}
  \vspace{-5mm}
\end{figure}

\noindent\textbf{Questions that need temporal reasoning.}
Every benchmark mixes questions answerable at a glance with questions that require relating observations across time (\cref{sec:reasoning_utility}).
We rank questions by how much a larger VLM, Qwen3.8-27B~\cite{qwen38}, gains from seeing the frames in order rather than shuffled, as one frame, or not at all (\appref{sec:app_temporal}).
Since shuffled frames show the same observations, a gain from order indicates that the answer depends on relating them rather than on seeing them, as in the cup-and-ball trick.
\Cref{fig:reasoning_need} shows that \method's gain grows with this need, reaching 8.1 points over Qwen2.5-VL on the top 5\%, versus 3.3 on all questions pooled.
This result adds valuable nuance to the simple six-benchmark average.

\noindent\textbf{Comparison with same-data SFT.}
Qwen2.5-VL-SFT is trained on the same SFT data~\cite{feng2025videor1} as \method.
It averages 61.2\%, 2.3 points below \method (SFT only), and is lower on every benchmark.
The same holds at 32 frames and with Qwen3-VL-8B~\cite{bai2025qwen3vl} as the backbone (\appsref{sec:app_32frames}{sec:qwen3}).
Learning from answers alone, \method thus reasons over video more effectively than a model trained on the same questions.

\vspace{-1mm}
\subsection{Ablations}
\label{sec:ablations}
\vspace{-1.5mm}
\begin{figure}[!t]
\centering
\includegraphics[width=\linewidth]{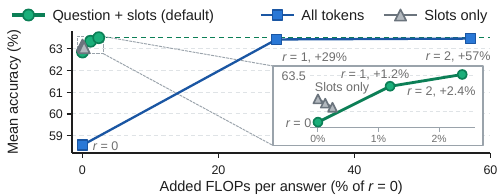}
\vspace{-7.5mm}
\caption{\textbf{Which tokens to loop.}
Six-benchmark mean accuracy at $r=0,1,2$ of separately trained SFT models, each with a different looping scheme, against added inference FLOPs relative to $r=0$.}
\label{fig:scope_ablation}
\vspace{-6mm}
\end{figure}

Ablations use 16 frames and report six-benchmark means.

\noindent\textbf{Which tokens to loop.}
\label{sec:depth_ablation}
\Cref{fig:scope_ablation} compares SFT models trained separately with the same data, each with a different looping scheme: the question and latent slots (ours), every input token including the video, as in looped vision-language models~\cite{qian2026loopvl,xu2026recursivevlm}, or the latent slots alone, as in latent-block recurrence~\cite{fan2026lotus}.
Our loop over the question and latent slots raises accuracy with two loops while adding only 2.4\% inference FLOPs.
Re-running every input token, including the video, reaches similar accuracy but adds 57\% FLOPs, whereas re-running only the slots leaves accuracy flat and ends 0.5 points lower.
This slots-only model is trained on the same data and is at least as accurate at $r=0$, so it also serves as a same-data latent-expansion control with the appended slots: the gain with depth comes from refining the question together with the slots, not from fine-tuning or the added slots.
What to loop therefore decides whether looping helps on video: the question must recur, but the video need not.

\begin{figure}[!t]
\centering
\begin{minipage}[t]{0.328\linewidth}
\centering
\includegraphics[width=\linewidth,trim=0 5pt 0 0,clip]{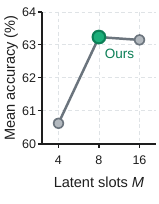}\\[-3pt]
{\footnotesize\textbf{(a)} Latent slots}
\end{minipage}\hfill
\begin{minipage}[t]{0.328\linewidth}
\centering
\includegraphics[width=\linewidth,trim=0 5pt 0 0,clip]{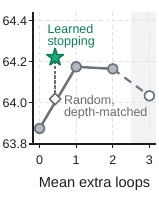}\\[-3pt]
{\footnotesize\textbf{(b)} Depth selection}
\end{minipage}\hfill
\begin{minipage}[t]{0.328\linewidth}
\centering
\includegraphics[width=\linewidth,trim=0 5pt 0 0,clip]{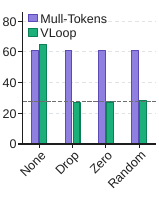}\\[-3pt]
{\footnotesize\textbf{(c)} Perturbed latents}
\end{minipage}
\vspace{-2mm}
\caption{\textbf{Latent slots, depth selection, and perturbing thinking latents.}
Six-benchmark mean accuracy.
(a)~SFT models with $M$ latent slots, averaged over $r=0,1,2$.
(b)~Exit policies after \alder; random selection matches the learned depth frequencies.
(c)~Mull-Tokens~\cite{ray2025mull} and \method on 3,000 questions with thinking latents dropped, zeroed, or resampled~\cite{zhang2026ablate,luo2026anchorsim}; dotted: chance.
}
\label{fig:slots_depth}
\vspace{-3mm}
\end{figure}

\noindent\textbf{Number of latent slots.}
Additional latent slots show diminishing returns (\cref{fig:slots_depth}a): accuracy rises from four to eight slots but not beyond, so we use $M=8$.

\begin{figure}[t]
  \centering
  \includegraphics[width=\linewidth]{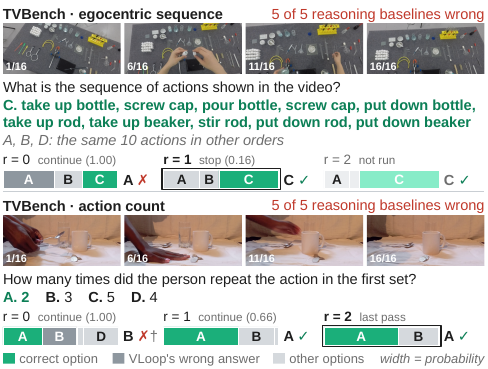}
  \vspace{-7mm}
  \caption{\textbf{Loops correct temporal answers that reasoning models get wrong.}
  Two questions that five video reasoning models~\cite{feng2025videor1,wang2025videorft,meng2025openo3,li2025lvr,ray2025mull} all answer wrongly.
  Bars: \method's option probabilities after $r$ loops; parentheses: probability of continuing; box: answering depth.
  Options shortened; $\dagger$\,near-tie.
  }
  \label{fig:qualitative}
  \vspace{-2mm}
\end{figure}

\noindent\textbf{What \alder learns.}
\label{sec:stopping_ablation}
Compared with the SFT model (\cref{fig:sft_depth_combined}), \alder raises accuracy at every exit, and the final model is more accurate than \method (SFT only) on all six benchmarks (\cref{tab:main_results}).
Its gain levels off after the first loop (\cref{fig:slots_depth}b), as in other looped models, where added recurrence brings diminishing returns and is harder to optimize~\cite{zhu2025ouro,geiping2025huginn}; we therefore do not train with more loops.
Because \alder also teaches the model when to reason further, continuing only where it expects an improvement, learned stopping is more accurate than fixed depth and random depth selection at the same mean depth.
\Cref{fig:qualitative} shows two temporal reasoning questions that all five reasoning baselines miss: ordering a sequence of actions and counting repetitions.
On both, the first loop changes \method's wrong answer at $r=0$ to the correct one, and \alder answers after one loop on the first question and after two on the second.

\noindent\textbf{Does the answer use the looped state?}
Following~\cite{zhang2026ablate,luo2026anchorsim}, we drop, zero, or resample thinking latents on 3,000 questions (\cref{fig:slots_depth}c): these tests leave Mull-Tokens' accuracy unchanged, whereas \method's brings it to chance.
Replacing the first loop's update with another question's changes 13\% of answers, against 3\% for removing it.

\vspace{-1mm}
\subsection{Inference Cost}
\label{sec:inference_efficiency}
\vspace{-2mm}
Each loop re-runs layers 10--20 over the question and slots only, reusing the cached video, so it adds 1.2\% inference FLOPs (\cref{fig:scope_ablation}).
As a result, \method is faster than representative video reasoning methods (text-CoT and latent expansion) on every benchmark while being more accurate (\cref{fig:reasoning_latency}).
On average, it answers about 5.0$\times$ faster than Video-R1 (text-CoT) and 1.9$\times$ faster than Mull-Tokens (latent-expansion); direct Qwen2.5-VL is faster but at the cost of accuracy.
\method thus makes video reasoning cheap: for only 0.2\,s more per question than direct answering, it improves on every benchmark, by up to 7.2 points on the temporal reasoning benchmark MVBench (\cref{tab:accuracy_latency,tab:main_results}).

\par\medskip
\noindent\begin{minipage}{\linewidth}
\centering
\captionof{table}{\textbf{Accuracy and latency.}
Six-benchmark mean accuracy (Avg., \%), smallest, median, and largest per-benchmark gain over direct answering (Min, Med., Max; pp), and mean latency per question (s), at 16 frames (H100, batch size one).
}
\vspace{-3mm}
\label{tab:accuracy_latency}
\footnotesize
\setlength{\tabcolsep}{4pt}
\renewcommand{\arraystretch}{1}
\resizebox{\ifdim\width>\linewidth\linewidth\else\width\fi}{!}{%
\begin{tabular}{llrrrrr}
\toprule
Method & Reasoning & Avg. & Min & Med. & Max & Latency \\
\midrule
Qwen2.5-VL~\cite{bai2025qwen25vl} & Direct & 60.3 & -- & -- & -- & \textbf{0.45} \\
Video-R1~\cite{feng2025videor1} & Text CoT & 60.9 & $-$0.4 & +0.1 & +2.5 & 3.24 \\
Mull-Tokens~\cite{ray2025mull} & Latent Exp. & 61.2 & $-$0.5 & +0.1 & +4.4 & 1.26 \\
\rowcolor{vloopgreen}
\textbf{\method (Ours)} & Latent Loop & \textbf{64.2} & \textbf{+0.8} & \textbf{+4.0} & \textbf{+7.2} & 0.65 \\
\bottomrule
\end{tabular}}
\par

\end{minipage}

\section{Conclusion}
\label{sec:conclusion}
For video-language models, rather than lengthening the reasoning sequence with textual rationales or latent tokens, this paper has argued for
\emph{deepening} the reasoning chain.
\method reads the video once and repeatedly refines the question and a small latent memory against the cached frames, so each additional loop costs little, and \alder teaches it when to think again.
Without requiring text rationales or prescribed latent targets, \method improves on its base model on every benchmark, with a mean gain more than four times the largest among text-CoT and latent-expansion methods with the same backbone, at a fraction of their latency.
Its gains are largest where reasoning matters most, on questions that require relating observations across time.
Our results offer a new view of how video models should reason: by thinking deeper about what they have already seen rather than by thinking longer.

{
    \small
    \bibliographystyle{ieeenat_fullname}
    \bibliography{main}
}

\clearpage
\appendix
\makeatletter
\setlength{\@dblfptop}{0pt}
\def\cvprsect#1{\cvprsection[#1]{\texorpdfstring{\hskip -1em.~}{}#1}}
\def\cvprsubsect#1{\cvprsubsection[#1]{\texorpdfstring{\hskip -1em.~}{}#1}}
\twocolumn[{%
  \pdfbookmark[0]{Appendix}{appendix}%
  {\noindent\Large\bfseries Appendix\par}%
  \vspace{1mm}%
  \@starttoc{toc}%
  \vspace{5mm}%
}]
\makeatother
\addtocontents{toc}{\protect\setcounter{tocdepth}{2}}

\section{Video Summary}
\label{sec:app_video}
We invite readers to watch the video summary on our project page, \url{https://vision.cs.utexas.edu/projects/vloop}, for an overview of \method, \alder, and the main results.
The page also animates the looped refinement and qualitative examples, showing loop by loop how \method works, how its answer changes, and when \alder decides to stop.

\section{Additional Results}
\label{sec:app_additional}
We extend the main results in four ways: more input frames, a second backbone, a control for the data that \alder trains on, and an ablation of the looped layers.

\subsection{Results at 32 Frames}
\label{sec:app_32frames}
\begin{table*}[!t]
\centering
\caption{\textbf{Main results at 32 frames.} Accuracy (\%) of 7B models at 32 frames, with the same score sources and notation as \cref{tab:main_results}; bold marks the best per column. \method models are the 16-frame-trained checkpoints evaluated with 32 input frames. Green: change from Qwen2.5-VL with direct answering at 32 frames.}
\vspace{-1mm}
\label{tab:main_results_32}
\providecolor{vloopgain}{HTML}{36745F}
\footnotesize
\setlength{\tabcolsep}{4pt}
\renewcommand{\arraystretch}{0.92}
\begin{tabular}{llrrrrrrr}
\toprule
 & & \multicolumn{3}{c}{Temporal reasoning} & \multicolumn{2}{c}{Complex \& causal reasoning} & \multicolumn{1}{c}{General} & \\
\cmidrule(lr){3-5}\cmidrule(lr){6-7}\cmidrule(lr){8-8}
Method & Reasoning & \multicolumn{1}{c}{MVBench} & \multicolumn{1}{c}{TempCompass} & \multicolumn{1}{c}{TVBench} & \multicolumn{1}{c}{VideoHolmes} & \multicolumn{1}{c}{NExT-QA} & \multicolumn{1}{c}{Video-MME} & \multicolumn{1}{c}{Avg.} \\
\midrule
Qwen2.5-VL~\cite{bai2025qwen25vl} & Direct & 64.2\phantom{\,{\tiny(+7.2)}} & 73.7\phantom{\,{\tiny(+7.2)}} & 53.1\phantom{\,{\tiny(+7.2)}} & 40.4\phantom{\,{\tiny(+7.2)}} & 81.5\phantom{\,{\tiny(+7.2)}} & 59.2\phantom{\,{\tiny(+7.2)}} & 62.0\phantom{\,{\tiny(+7.2)}} \\
Video-R1~\cite{feng2025videor1} & Text CoT & 63.9\phantom{\,{\tiny(+7.2)}} & 73.2\phantom{\,{\tiny(+7.2)}} & 52.4\phantom{\,{\tiny(+7.2)}} & 42.2\phantom{\,{\tiny(+7.2)}} & 79.7\phantom{\,{\tiny(+7.2)}} & 59.3\phantom{\,{\tiny(+7.2)}} & 61.8\phantom{\,{\tiny(+7.2)}} \\
VideoRFT~\cite{wang2025videorft} & Text CoT & 62.1\phantom{\,{\tiny(+7.2)}} & 73.7\phantom{\,{\tiny(+7.2)}} & 55.3\phantom{\,{\tiny(+7.2)}} & 41.3\phantom{\,{\tiny(+7.2)}} & 77.7\phantom{\,{\tiny(+7.2)}} & 59.8\phantom{\,{\tiny(+7.2)}} & 61.6\phantom{\,{\tiny(+7.2)}} \\
Video-VER~\cite{luo2025drift} & Text CoT & 64.1\phantom{\,{\tiny(+7.2)}} & 74.0\phantom{\,{\tiny(+7.2)}} & 52.8\phantom{\,{\tiny(+7.2)}} & --\phantom{\,{\tiny(+7.2)}} & --\phantom{\,{\tiny(+7.2)}} & 59.3\phantom{\,{\tiny(+7.2)}} & --\phantom{\,{\tiny(+7.2)}} \\
Open-o3-Video~\cite{meng2025openo3} & Text CoT & 66.0\phantom{\,{\tiny(+7.2)}} & 71.8\phantom{\,{\tiny(+7.2)}} & 56.8\phantom{\,{\tiny(+7.2)}} & 41.2\phantom{\,{\tiny(+7.2)}} & 81.2\phantom{\,{\tiny(+7.2)}} & 61.3\phantom{\,{\tiny(+7.2)}} & 63.0\phantom{\,{\tiny(+7.2)}} \\
VIOS~\cite{dugad2026vios} & Text CoT & 65.5\phantom{\,{\tiny(+7.2)}} & 72.8\phantom{\,{\tiny(+7.2)}} & 52.2\phantom{\,{\tiny(+7.2)}} & --\phantom{\,{\tiny(+7.2)}} & --\phantom{\,{\tiny(+7.2)}} & 59.6\phantom{\,{\tiny(+7.2)}} & --\phantom{\,{\tiny(+7.2)}} \\
LVR~\cite{li2025lvr} & Latent Expansion & 65.1\phantom{\,{\tiny(+7.2)}} & 72.0\phantom{\,{\tiny(+7.2)}} & 53.0\phantom{\,{\tiny(+7.2)}} & 34.8\phantom{\,{\tiny(+7.2)}} & 81.5\phantom{\,{\tiny(+7.2)}} & 60.5\phantom{\,{\tiny(+7.2)}} & 61.1\phantom{\,{\tiny(+7.2)}} \\
Mull-Tokens~\cite{ray2025mull} & Latent Expansion & 68.5\phantom{\,{\tiny(+7.2)}} & 73.6\phantom{\,{\tiny(+7.2)}} & 53.6\phantom{\,{\tiny(+7.2)}} & 40.3\phantom{\,{\tiny(+7.2)}} & 79.9\phantom{\,{\tiny(+7.2)}} & 59.5\phantom{\,{\tiny(+7.2)}} & 62.6\phantom{\,{\tiny(+7.2)}} \\
STORM~\cite{liang2026torm} & Latent Expansion & 61.1\phantom{\,{\tiny(+7.2)}} & 74.3\phantom{\,{\tiny(+7.2)}} & --\phantom{\,{\tiny(+7.2)}} & 37.8\phantom{\,{\tiny(+7.2)}} & --\phantom{\,{\tiny(+7.2)}} & 61.0\phantom{\,{\tiny(+7.2)}} & --\phantom{\,{\tiny(+7.2)}} \\
VTI-CoT~\cite{zhang2026vticot} & Latent Expansion & 65.9\phantom{\,{\tiny(+7.2)}} & 74.5\phantom{\,{\tiny(+7.2)}} & --\phantom{\,{\tiny(+7.2)}} & --\phantom{\,{\tiny(+7.2)}} & --\phantom{\,{\tiny(+7.2)}} & 59.6\phantom{\,{\tiny(+7.2)}} & --\phantom{\,{\tiny(+7.2)}} \\
VideoLatent~\cite{hu2026videolatent} & Latent Expansion & 69.4\phantom{\,{\tiny(+7.2)}} & 74.1\phantom{\,{\tiny(+7.2)}} & --\phantom{\,{\tiny(+7.2)}} & --\phantom{\,{\tiny(+7.2)}} & --\phantom{\,{\tiny(+7.2)}} & 61.4\phantom{\,{\tiny(+7.2)}} & --\phantom{\,{\tiny(+7.2)}} \\
\cmidrule(lr){1-9}
Qwen2.5-VL-SFT$^\dagger$ & Direct & 67.3\phantom{\,{\tiny(+7.2)}} & 74.6\phantom{\,{\tiny(+7.2)}} & 54.3\phantom{\,{\tiny(+7.2)}} & 41.5\phantom{\,{\tiny(+7.2)}} & 80.6\phantom{\,{\tiny(+7.2)}} & 59.3\phantom{\,{\tiny(+7.2)}} & 62.9\phantom{\,{\tiny(+7.2)}} \\
\rowcolor{vloopgreen}
\textbf{\method (SFT only)} & Latent Loop & 69.8\phantom{\,{\tiny(+7.2)}} & 75.0\phantom{\,{\tiny(+7.2)}} & 58.2\phantom{\,{\tiny(+7.2)}} & 42.2\phantom{\,{\tiny(+7.2)}} & 81.7\phantom{\,{\tiny(+7.2)}} & \textbf{62.2}\phantom{\,{\tiny(+7.2)}} & 64.8\phantom{\,{\tiny(+7.2)}} \\
\rowcolor{vloopgreen}
\textbf{\method (Ours)} & Latent Loop & \textbf{70.4}\rlap{\,{\tiny\color{vloopgain}(+6.2)}}\phantom{\,{\tiny(+7.2)}} & \textbf{76.1}\rlap{\,{\tiny\color{vloopgain}(+2.4)}}\phantom{\,{\tiny(+7.2)}} & \textbf{59.2}\rlap{\,{\tiny\color{vloopgain}(+6.1)}}\phantom{\,{\tiny(+7.2)}} & \textbf{42.8}\rlap{\,{\tiny\color{vloopgain}(+2.4)}}\phantom{\,{\tiny(+7.2)}} & \textbf{81.9}\rlap{\,{\tiny\color{vloopgain}(+0.4)}}\phantom{\,{\tiny(+7.2)}} & 61.6\rlap{\,{\tiny\color{vloopgain}(+2.4)}}\phantom{\,{\tiny(+7.2)}} & \textbf{65.3}\rlap{\,{\tiny\color{vloopgain}(+3.3)}}\phantom{\,{\tiny(+7.2)}} \\
\bottomrule
\end{tabular}
\end{table*}

\Cref{tab:main_results_32} repeats the main comparison at 32 frames, adding VIOS, STORM, and VTI-CoT, which report only 32-frame results and release no checkpoints.
The findings of \cref{sec:reasoning_comparison} hold at 32 frames.
\method is more accurate than every baseline on all six benchmarks and achieves the highest mean, 65.3\%, a gain of 3.3 points over direct Qwen2.5-VL, against at most 1.0 for the reasoning baselines with all six scores.
The margins are again largest on the temporal reasoning benchmarks: on TVBench, \method is 2.4 points above the best baseline, versus 0.4 and 0.2 on NExT-QA and Video-MME.
The final model is more accurate than \method (SFT only) on five of six benchmarks.
Qwen2.5-VL-SFT averages 62.9\%, 1.9 points below \method (SFT only), and is lower on each benchmark.

\subsection{Qwen3-VL Backbone}
\label{sec:qwen3}
\begin{table*}[!t]
\centering
\caption{\textbf{\method on Qwen3-VL-8B.} Accuracy (\%) at 16 and 32 frames with Qwen3-VL-8B-Instruct as the backbone of every row; all rows use identical frames within each frame budget. 32-frame rows evaluate the 16-frame-trained checkpoints with 32 input frames. $^\dagger$: SFT on the same training data as \method. Green rows are ours: SFT only uses $r=2$; the final model (Ours) adds \alder and uses adaptive depth. Qwen3-VL answers directly, and its answer letter is credited however it is formatted (e.g., \texttt{<A> </A>}). Green: change from Qwen3-VL with direct answering at the same frame budget. Bold marks the best per column within each frame budget.}
\vspace{-1mm}
\label{tab:qwen3_results}
\providecolor{vloopgain}{HTML}{36745F}
\footnotesize
\setlength{\tabcolsep}{4pt}
\renewcommand{\arraystretch}{0.92}
\begin{tabular}{llrrrrrrr}
\toprule
 & & \multicolumn{3}{c}{Temporal reasoning} & \multicolumn{2}{c}{Complex \& causal reasoning} & \multicolumn{1}{c}{General} & \\
\cmidrule(lr){3-5}\cmidrule(lr){6-7}\cmidrule(lr){8-8}
Method & Reasoning & \multicolumn{1}{c}{MVBench} & \multicolumn{1}{c}{TempCompass} & \multicolumn{1}{c}{TVBench} & \multicolumn{1}{c}{VideoHolmes} & \multicolumn{1}{c}{NExT-QA} & \multicolumn{1}{c}{Video-MME} & \multicolumn{1}{c}{Avg.} \\
\midrule
\multicolumn{9}{l}{\textit{16 frames}} \\
Qwen3-VL~\cite{bai2025qwen3vl} & Direct & 66.2\phantom{\,{\tiny(+7.2)}} & 76.5\phantom{\,{\tiny(+7.2)}} & 56.8\phantom{\,{\tiny(+7.2)}} & 40.1\phantom{\,{\tiny(+7.2)}} & 81.0\phantom{\,{\tiny(+7.2)}} & 60.9\phantom{\,{\tiny(+7.2)}} & 63.6\phantom{\,{\tiny(+7.2)}} \\
Qwen3-VL-SFT$^\dagger$ & Direct & 65.8\phantom{\,{\tiny(+7.2)}} & 73.7\phantom{\,{\tiny(+7.2)}} & 53.7\phantom{\,{\tiny(+7.2)}} & 42.2\phantom{\,{\tiny(+7.2)}} & 79.1\phantom{\,{\tiny(+7.2)}} & 59.7\phantom{\,{\tiny(+7.2)}} & 62.3\phantom{\,{\tiny(+7.2)}} \\
\rowcolor{vloopgreen}
\textbf{\method (SFT only)} & Latent Loop & 69.7\rlap{\,{\tiny\color{vloopgain}(+3.5)}}\phantom{\,{\tiny(+7.2)}} & 76.0\rlap{\,{\tiny\color{vloopgain}(-0.5)}}\phantom{\,{\tiny(+7.2)}} & 59.2\rlap{\,{\tiny\color{vloopgain}(+2.4)}}\phantom{\,{\tiny(+7.2)}} & 46.5\rlap{\,{\tiny\color{vloopgain}(+6.4)}}\phantom{\,{\tiny(+7.2)}} & 82.0\rlap{\,{\tiny\color{vloopgain}(+1.0)}}\phantom{\,{\tiny(+7.2)}} & 60.6\rlap{\,{\tiny\color{vloopgain}(-0.3)}}\phantom{\,{\tiny(+7.2)}} & 65.7\rlap{\,{\tiny\color{vloopgain}(+2.1)}}\phantom{\,{\tiny(+7.2)}} \\
\rowcolor{vloopgreen}
\textbf{\method (Ours)} & Latent Loop & \textbf{70.0}\rlap{\,{\tiny\color{vloopgain}(+3.8)}}\phantom{\,{\tiny(+7.2)}} & \textbf{79.1}\rlap{\,{\tiny\color{vloopgain}(+2.6)}}\phantom{\,{\tiny(+7.2)}} & \textbf{61.9}\rlap{\,{\tiny\color{vloopgain}(+5.1)}}\phantom{\,{\tiny(+7.2)}} & \textbf{47.4}\rlap{\,{\tiny\color{vloopgain}(+7.3)}}\phantom{\,{\tiny(+7.2)}} & \textbf{82.9}\rlap{\,{\tiny\color{vloopgain}(+1.9)}}\phantom{\,{\tiny(+7.2)}} & \textbf{62.1}\rlap{\,{\tiny\color{vloopgain}(+1.2)}}\phantom{\,{\tiny(+7.2)}} & \textbf{67.2}\rlap{\,{\tiny\color{vloopgain}(+3.6)}}\phantom{\,{\tiny(+7.2)}} \\
\midrule
\multicolumn{9}{l}{\textit{32 frames}} \\
Qwen3-VL~\cite{bai2025qwen3vl} & Direct & 68.3\phantom{\,{\tiny(+7.2)}} & 76.9\phantom{\,{\tiny(+7.2)}} & 59.5\phantom{\,{\tiny(+7.2)}} & 40.6\phantom{\,{\tiny(+7.2)}} & 82.7\phantom{\,{\tiny(+7.2)}} & 64.5\phantom{\,{\tiny(+7.2)}} & 65.4\phantom{\,{\tiny(+7.2)}} \\
Qwen3-VL-SFT$^\dagger$ & Direct & 67.4\phantom{\,{\tiny(+7.2)}} & 74.8\phantom{\,{\tiny(+7.2)}} & 56.5\phantom{\,{\tiny(+7.2)}} & 43.7\phantom{\,{\tiny(+7.2)}} & 80.3\phantom{\,{\tiny(+7.2)}} & 62.8\phantom{\,{\tiny(+7.2)}} & 64.2\phantom{\,{\tiny(+7.2)}} \\
\rowcolor{vloopgreen}
\textbf{\method (SFT only)} & Latent Loop & 70.7\rlap{\,{\tiny\color{vloopgain}(+2.4)}}\phantom{\,{\tiny(+7.2)}} & 76.1\rlap{\,{\tiny\color{vloopgain}(-0.8)}}\phantom{\,{\tiny(+7.2)}} & 60.3\rlap{\,{\tiny\color{vloopgain}(+0.8)}}\phantom{\,{\tiny(+7.2)}} & 48.8\rlap{\,{\tiny\color{vloopgain}(+8.2)}}\phantom{\,{\tiny(+7.2)}} & 82.7\rlap{\,{\tiny\color{vloopgain}(+0.0)}}\phantom{\,{\tiny(+7.2)}} & \textbf{64.9}\rlap{\,{\tiny\color{vloopgain}(+0.4)}}\phantom{\,{\tiny(+7.2)}} & 67.2\rlap{\,{\tiny\color{vloopgain}(+1.8)}}\phantom{\,{\tiny(+7.2)}} \\
\rowcolor{vloopgreen}
\textbf{\method (Ours)} & Latent Loop & \textbf{71.3}\rlap{\,{\tiny\color{vloopgain}(+3.0)}}\phantom{\,{\tiny(+7.2)}} & \textbf{79.4}\rlap{\,{\tiny\color{vloopgain}(+2.5)}}\phantom{\,{\tiny(+7.2)}} & \textbf{64.3}\rlap{\,{\tiny\color{vloopgain}(+4.8)}}\phantom{\,{\tiny(+7.2)}} & \textbf{49.9}\rlap{\,{\tiny\color{vloopgain}(+9.3)}}\phantom{\,{\tiny(+7.2)}} & \textbf{83.8}\rlap{\,{\tiny\color{vloopgain}(+1.1)}}\phantom{\,{\tiny(+7.2)}} & 64.4\rlap{\,{\tiny\color{vloopgain}(-0.1)}}\phantom{\,{\tiny(+7.2)}} & \textbf{68.8}\rlap{\,{\tiny\color{vloopgain}(+3.4)}}\phantom{\,{\tiny(+7.2)}} \\
\bottomrule
\end{tabular}
\vspace{-3mm}
\end{table*}

\Cref{tab:qwen3_results} repeats the main comparison with Qwen3-VL-8B-Instruct~\cite{bai2025qwen3vl} as the backbone.
\method improves on direct Qwen3-VL by 3.6 points at 16 frames and 3.4 at 32, and \alder adds 1.5 and 1.6 points over \method (SFT only), more than with Qwen2.5-VL (0.7 and 0.5).
Qwen3-VL-SFT, trained with the same SFT data as \method, is 3.4 and 3.0 points below \method (SFT only) and lower on every benchmark, so the gain comes from the loop rather than from the data.

\noindent\textbf{Looped layers.}
Qwen3-VL's decoder has 36 layers rather than 28, so we locate the recurrent block with a probe of the pretrained backbone.
At each decoder layer, we measure how much the question states change and how much they depend on the video, the latter by replacing the video with another.
On Qwen2.5-VL, the per-layer change settles at layer 10 and rises again at layer 21, and the dependence on the video peaks in between.
The probe thus marks layers 10--20, the same layers that prior work~\cite{yu2025multimodallayers} identifies as the reasoning layers of Qwen2-VL, which has the same 28-layer decoder.
On Qwen3-VL, the same landmarks fall at layers 13 and 22, so the recurrent block comprises layers 13--21 (zero-indexed).

\noindent\textbf{Training.}
Frames are resized with Qwen3-VL's own video preprocessing, with at most 401,408 pixels per frame.
SFT uses the same data and recipe as for Qwen2.5-VL (\cref{sec:setup}).
For \alder, the RL data are reselected from the same candidates with the rule in \appref{sec:app_rl_data}, profiled by the Qwen3-VL SFT model, giving 14,896 questions and one epoch of 621 updates; the RL recipe is unchanged.

\subsection{Supervised Training on the RL Data}
\label{sec:app_rl_data_sft}
\alder adds 15,237 training questions (\appref{sec:app_rl_data}).
To separate their effect from that of \alder, we continue training \method (SFT only) on the same questions for one epoch, with \cref{eq:answer_sft} and $R_{\mathrm{SFT}}=2$.

\begin{figure}[t]
\centering
\includegraphics[width=\linewidth]{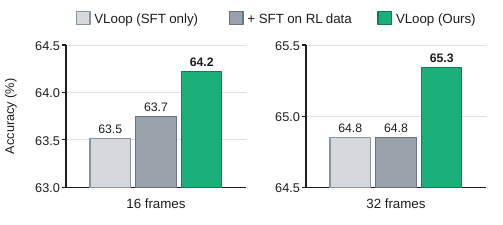}
\vspace{-9mm}
\caption{\textbf{Supervised training on the RL data.} Six-benchmark mean accuracy (\%) at 16 and 32 frames; the SFT models are evaluated at $r=2$. \method (SFT only) and \method (Ours) are the rows of \cref{tab:main_results,tab:main_results_32}. Axes do not start at zero.}
\label{fig:rl_data_sft}
\par\vspace{1mm}
\includegraphics[width=\linewidth]{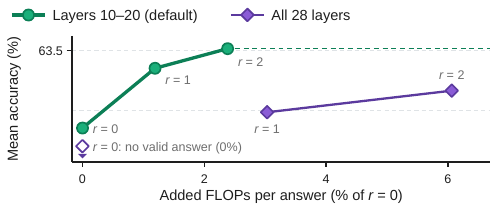}
\vspace{-7mm}
\caption{\textbf{Which layers to loop.} Six-benchmark mean accuracy against added inference FLOPs relative to $r=0$, for separately trained SFT models that loop the question and latent slots through layers 10--20 or through all 28 layers. The all-layer model gives no valid answer at $r=0$.
}

\label{fig:loop_layers}
\end{figure}

The additional training raises the six-benchmark mean by 0.2 points at 16 frames and 0.0 at 32, whereas \alder raises it by 0.7 and 0.5 (\cref{fig:rl_data_sft}).
The gain of \method over \method (SFT only) therefore comes from \alder, not from the additional questions.

\subsection{Which Layers to Loop}
\label{sec:app_loop_layers}
\method loops decoder layers 10--20 of Qwen2.5-VL~\cite{bai2025qwen25vl}, following prior work that places reasoning in these layers~\cite{yu2025multimodallayers}.
\Cref{fig:loop_layers} compares \method (SFT only) with a model trained identically, except that the question and latent slots loop through all 28 layers.
Over layers 10--20, accuracy rises with every loop, and each loop adds 1.2\% FLOPs.
Over all layers, accuracy also rises with a second loop, but one loop over layers 10--20 reaches higher accuracy than two loops over all layers, at one-fifth of the added FLOPs.
The all-layer model gives no valid answer at $r=0$, so it cannot stop without a loop.
Re-running the early and late layers therefore adds cost but not accuracy.

\section{Questions That Need Temporal Reasoning}
\label{sec:app_temporal}
This section details the question labels behind \cref{fig:reasoning_need} and reports the gains on the labeled questions.

\begin{figure*}[t]
  \centering
  \includegraphics[width=\linewidth]{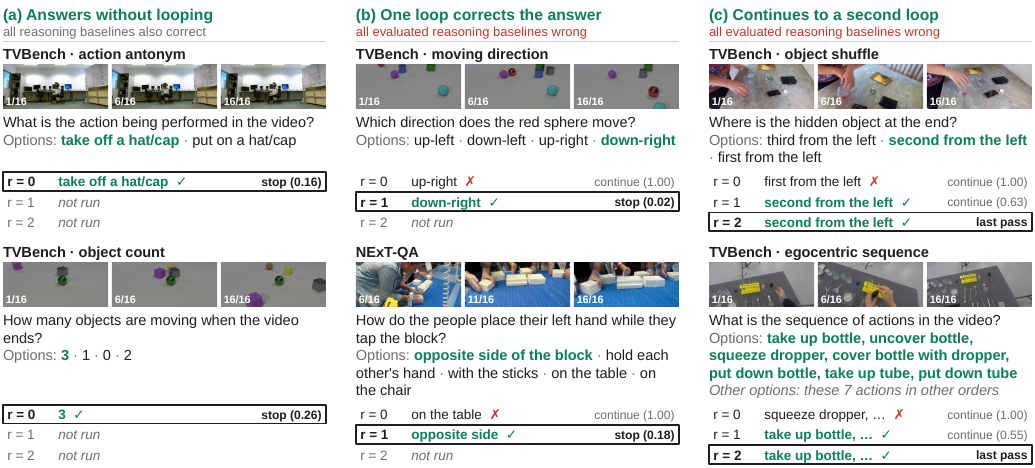}
  \vspace{-5mm}
  \caption{\textbf{Adaptive depth in practice.}
  Six questions answered by \method.
  Each line gives the model's answer after $r$ loops and its stopping decision, with the probability of continuing in parentheses; the box marks the answering depth, and later loops are not run.
  The correct option is in green.
  The reasoning baselines are those we evaluate on each benchmark.
  Questions and options are shortened.}
  \label{fig:qualitative_app}
\end{figure*}

\begin{figure*}[t]
  \centering
  \includegraphics[width=\linewidth]{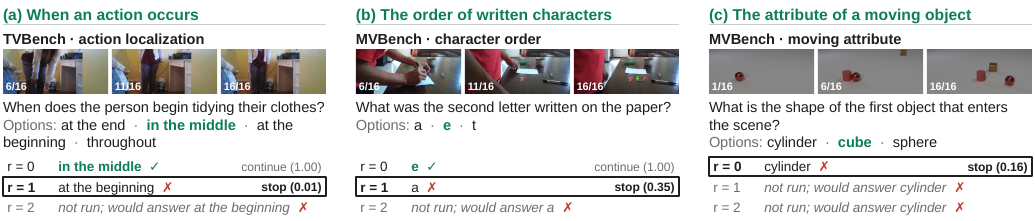}
  \vspace{-5mm}
  \caption{\textbf{Failure cases.}
  One failure of \method on each kind of question that \cref{sec:depth_diagnostic} finds to be decided by a precise detail, in the format of \cref{fig:qualitative_app}.
  In (a) and (b), a loop overturns a correct first answer; in (c), the answer is wrong at every depth.
  For loops that are not run, the answer they would give is shown in gray.
  Questions and options are shortened.}
  \label{fig:qualitative_failure}
\end{figure*}

\noindent\textbf{Probe model and inputs.}
The labels come from Qwen3.8-27B~\cite{qwen38}, a newer model than the Qwen2.5-VL backbone of all compared methods, used without training.
It answers all 27,166 questions of the six benchmarks with the same 16 frames as \method, under four inputs: the question and options alone; the middle frame as one image; the 16 frames as a video in a fixed random order; and the 16 frames as a video in their original order.
With thinking disabled, we take its probabilities of the offered option letters.
The probe's accuracy is 47.1\% from the question alone, 55.4\% from one frame, 62.4\% from shuffled frames, and 71.1\% from frames in order.

\noindent\textbf{Label and ranking.}
For a question with $K$ options, let $p_c$ be the probe's probability of the correct option under input $c$ and $s_c=(p_c-1/K)/(1-1/K)$, so that chance gives 0 and certainty gives 1.
We define $\Delta = s_{\text{ordered}} - \max(s_{\text{text}}, s_{\text{frame}}, s_{\text{shuffled}})$.
A question is labeled as needing temporal reasoning if $\Delta>0$, that is, if the frames in order raise the probability of the correct answer above all three other inputs; this holds for 10,005 of the 27,166 questions (36.8\%).
\Cref{fig:reasoning_need} ranks questions by decreasing $\Delta$.
Gains are pooled over questions rather than averaged over benchmarks, so the gain over Qwen2.5-VL on all questions (3.3 points) differs from the benchmark-mean gain in \cref{tab:main_results} (3.9).
Confidence intervals come from a paired bootstrap over source videos (5,000 resamples), which resamples questions on the same video together.

\noindent\textbf{Results.}
Pooled over the six benchmarks, \method gains 4.8 points over Qwen2.5-VL on questions that need temporal reasoning and 2.4 on the rest; over \method (SFT only), it gains 2.4 points and $-0.2$.
The differences between the groups, 2.5 [1.5, 3.6] and 2.6 [1.9, 3.3] points, are both significant.
They hold within benchmarks: the gain over \method (SFT only) is larger on labeled questions in all six, and the gain over Qwen2.5-VL in five.
They do not come from more room to improve, since all three models are more accurate on the labeled questions than on the rest (\method: 78.8\% vs.\ 66.3\%).
The label also agrees with the benchmarks' design, marking 42--49\% of the questions in the three temporal reasoning benchmarks and 21--28\% in the other three.
Under a stricter label, the 1,457 questions that the probe answers correctly only from the frames in order, the gains grow to 10.3 points [7.3, 13.5] over Qwen2.5-VL and 6.5 [4.6, 8.5] over \method (SFT only).
In short, \method's gain over direct answering doubles on the questions that need temporal reasoning, and the gain from \alder comes from these questions alone.

\section{Qualitative Examples}
\label{sec:app_qualitative}
\begin{table*}[t]
\begin{minipage}[t]{0.60\linewidth}
\centering
\caption{\textbf{Latency per benchmark.} Mean seconds per question at 16 frames, on 30 fixed questions per benchmark with three repetitions on H100 GPUs (batch size one, BF16). All methods receive the same cached frames. Timings include visual encoding, reasoning, stopping decisions, and answer generation, and exclude input preparation. Released models keep their own inference backends and prompts.}
\vspace{-1mm}
\label{tab:latency_all}
\footnotesize
\setlength{\tabcolsep}{6pt}
\begin{tabular}{lrrrr}
\toprule
Benchmark & Qwen2.5-VL & Video-R1 & Mull-Tokens & \method \\
\midrule
MVBench & 0.42 & 3.15 & 1.10 & 0.64 \\
TempCompass & 0.46 & 3.21 & 1.37 & 0.64 \\
TVBench & 0.44 & 3.10 & 1.18 & 0.67 \\
VideoHolmes & 0.47 & 3.45 & 1.39 & 0.68 \\
NExT-QA & 0.41 & 3.22 & 1.09 & 0.57 \\
Video-MME & 0.48 & 3.32 & 1.41 & 0.67 \\
\bottomrule
\end{tabular}
\end{minipage}\hfill
\begin{minipage}[t]{0.34\linewidth}
\centering
\caption{\textbf{RL training data by source family.}}
\label{tab:rl_sources}
\footnotesize
\setlength{\tabcolsep}{6pt}
\begin{tabular}{lr}
\toprule
Source & Questions \\
\midrule
STAR & 2,800 \\
CLEVRER & 2,800 \\
RTime & 2,800 \\
SEED-Bench-R1 & 2,779 \\
Video-R1 subset & 598 \\
VideoKR & 2,763 \\
Perception Test & 697 \\
\midrule
Total & 15,237 \\
\bottomrule
\end{tabular}
\end{minipage}
\vspace{-2.5mm}
\end{table*}

\Cref{fig:qualitative_app} shows how \method spends its loops on six questions.
In (a), the model stops at $r=0$ and answers without looping; every reasoning baseline also answers these questions correctly.
In (b) and (c), the first pass gives a wrong answer, as does every reasoning baseline we evaluate.
After one loop the answer is correct, and the model then stops (b) or continues to a second loop, which keeps the answer (c).
These examples illustrate the behavior \alder trains for: looping only where it can change the answer.

\noindent\textbf{Failure cases.}
\Cref{sec:depth_diagnostic} finds that looping lowers accuracy on questions decided by a precise detail: when an action occurs, the order of written characters, or the attribute of a moving object.
\Cref{fig:qualitative_failure} shows one failure of \method on each.
In (a) and (b), the first pass is correct and a loop overturns the answer; in (c), the model is wrong at every depth.
Refinement reuses the evidence already encoded, so it does not recover a detail the first pass missed, and further updates can drift from it.
These failures concern what the video encoding captures, not how long the model reasons, so they call for richer visual evidence, such as more input frames, and not for more loops.

\section{Inference Cost}
\label{sec:app_cost}

\noindent\textbf{FLOPs.}
We count inference FLOPs per answer analytically, from the backbone's dimensions and each question's token counts.
At $r=0$, the count covers the vision encoder, one pass of the video and question through all 28 decoder layers, and a 10-token answer.
A loop re-runs layers 10--20, 11 of the 28, over the looped positions only, reusing the video cache.
For \method these are the question and slots, about 80 of the roughly 1,950 positions in the prompt, so one loop adds 1.2\% of the $r=0$ FLOPs and two loops add 2.4\%.
Looping every token re-runs the same layers over the whole prompt, which costs 24 times more per loop and adds 57\% for two loops (\cref{fig:scope_ablation}).

\noindent\textbf{Latency.}
\Cref{tab:latency_all} gives the per-benchmark latencies averaged in \cref{tab:accuracy_latency}.
Latency rises by more than FLOPs because the two measure different things.
Most FLOPs are spent on the video tokens, which the GPU processes in one parallel pass, whereas latency at batch size one grows with the number of sequential decoder calls: a call over a few tokens needs little arithmetic but still runs every layer in turn.
Text CoT is slow for this reason, since it decodes its rationale one token at a time.
\method adds only a few short calls, for its loops, its stopping decisions, and its answer, so it adds 0.2\,s to direct answering where Video-R1 adds 2.8\,s.

\section{Architecture and SFT Details}
\label{sec:looped_implementation}

We use Qwen2.5-VL-7B~\cite{bai2025qwen25vl}, whose decoder has 28 layers.
The recurrent block $F_\theta$ comprises layers 10--20 (zero-indexed), following layer-wise probes that place reasoning in the middle layers~\cite{yu2025multimodallayers}; the upper block $G_\theta$ includes the final decoder normalization.
The $M=8$ slots share a seed initialized from the latent-start embedding and optimized as a separate parameter, and each slot occupies a distinct sequence position.
The post-video prompt $\mathbf{E}_q$ includes question text, answer options, and formatting tokens.
Question and slot keys and values are recomputed at every pass, using the same weights and sequence positions.
SFT uses 80,614 questions for two epochs (2,443 updates).
Both SFT and \alder use AdamW with weight decay 0.01, gradient clipping at norm 1, and a learning rate that warms up linearly to $10^{-6}$ over the first 5\% of updates and then decays linearly to zero.

\section{\alder Implementation Details}
\label{sec:ald_implementation}

\alder is initialized from the SFT checkpoint analyzed in \cref{sec:depth_diagnostic}.
The answer and stopping objectives use the same training questions but have separate optimizers and gradients.

\subsection{All-Depth Answer Objective}
\label{sec:ald_answer_details}

For input $x=(V,q)$, let $\pi_\theta^{(r)}$ denote the answer distribution at exit $r$.
At each exit we sample $K$ answers $y_i^{(r)}$ from the behavior policy $\pi_{\theta_{\mathrm{old}}}^{(r)}$ and compute the leave-one-out advantages in \cref{eq:depth_advantage}.
The clipped objective is
\begin{equation}
    \mathcal{L}_{\mathrm{GRPO}}
    =-\mathbb{E}\!\left[\frac{1}{(R+1)K}
        \sum_{r=0}^{R}\sum_{i=1}^{K}\sum_{t=1}^{T_i^{(r)}}
        \ell_{i,t}^{(r)}\right],
    \label{eq:all_depth_grpo}
\end{equation}
where $\ell_{i,t}^{(r)}=\ell(\rho_{i,t}^{(r)},A_i^{(r)})$, $\ell(\rho,A)=\min\{\rho A,\operatorname{clip}(\rho,1-\epsilon,1+\epsilon)A\}$, $\rho_{i,t}^{(r)}$ is the current-to-behavior probability ratio of the sampled token given $x$, depth $r$, and the preceding answer tokens, and $T_i^{(r)}$ is the response length.
We use $K=8$ sampled answers per exit, clipping threshold $\epsilon=0.2$, and $\beta_{\mathrm{ref}}=\beta_{\mathrm{perm}}=0.05$.

\noindent\textbf{Answer decoding.}
At each depth, sampled and greedy answers stop at the first valid boxed answer matching an offered option, EOS, or a 128-token limit.
Invalid answers and responses that exhaust the token budget receive zero reward.

\noindent\textbf{Reference and permutation-consistency losses.}
$\mathcal{L}_{\mathrm{ref}}$ is the KL from the frozen SFT distribution over the offered answers to the current model at $r=0$.
For permutation consistency, the distribution under permuted answer choices is mapped back to the original choice identities and detached; $\mathcal{L}_{\mathrm{perm}}$ sums the KL from this teacher to the current distribution over all exits.

\subsection{Stopping Objective}
\label{sec:ald_stopping_details}

\noindent\textbf{Stopping features.}
A single head is shared across the two nonterminal exits, with nine inputs, a 16-unit tanh hidden layer, and a scalar continuation logit.
Three inputs describe the current answer distribution: the maximum offered-answer probability, the normalized entropy, and the log-probability margin between the two highest-scoring options.
Five describe its change from the preceding pass and are zero at $r=0$: whether the preferred option changed, the changes in the three inputs above, and the normalized Jensen--Shannon divergence between the two distributions.
The ninth is a depth flag.

\noindent\textbf{Online updates.}
Every 16 answer-model updates, we recompute the greedy outcomes and stopping features at all depths for the latest 64 batches, with the backbone held fixed, and take 16 AdamW steps on the head (learning rate $0.01$) to maximize $J_\phi$ in \cref{eq:stopping_return}.
We add a prior toward equal stop/continue probabilities, $D_{\mathrm{KL}}(\operatorname{Bern}(\tfrac12)\,\|\,\operatorname{Bern}(h_r))$ averaged over the nonterminal exits, which discourages prematurely deterministic decisions; its weight decays from $10^{-2}$ to $10^{-4}$ on a cosine schedule.

\subsection{RL Training Data}
\label{sec:app_rl_data}

\noindent\textbf{Sources.}
The RL data contain multiple-choice questions from STAR~\cite{wu2021star}, CLEVRER~\cite{yi2019clevrer}, SEED-Bench-R1~\cite{chen2025seedbenchr1}, VideoKR~\cite{fu2026videokr}, and Perception Test~\cite{patraucean2023perception}, including action-order questions derived from Perception Test annotations.
They also include Video-R1-derived questions~\cite{feng2025videor1} and questions on RTime videos~\cite{du2024reversed}, whose forward and time-reversed versions count as one source family.
All questions come from training splits, and a check of video identifiers finds none of their videos in the six evaluation benchmarks.

\noindent\textbf{Selection.}
Candidate questions are profiled with a \method SFT model; we keep those whose gold-answer probability after two loops lies between 0.1 and 0.9.
To balance sources, each source family is capped at 2,800 questions, yielding 15,237 (\cref{tab:rl_sources}).

\section{Evaluation Protocol}
\label{sec:results_protocol}

\noindent\textbf{Evaluation populations and score sources.}
Our evaluations use the complete benchmarks: MVBench (4,000 questions), TempCompass (7,540, in the multiple-choice version of Video-R1~\cite{feng2025videor1}), TVBench (2,525), VideoHolmes (1,837), NExT-QA (8,564), and Video-MME (2,700).
For baselines, published scores take precedence: we use the original paper where it reports a result, and otherwise the results reported by VideoLatent, DyLaR, and Video-VER~\cite{hu2026videolatent,xia2026dylar,luo2025drift}.
We evaluate only the remaining benchmarks of baselines with public code and checkpoints.

\noindent\textbf{Baseline inference.}
The reasoning baselines we evaluate are Video-R1, Mull-Tokens, Open-o3-Video, VideoRFT, and LVR.
They receive the same sampled frames and pixel budget as \method and run their released inference code and prompts with greedy decoding.
LVR is an image model, so we give it the sampled frames as one video with the same visual tokens as \method and use the number of latent steps (16) that best reproduces its published MVBench accuracy.
In \cref{fig:reasoning_latency}, the MVBench panel plots our evaluations of Video-R1 and Mull-Tokens on these frames, whereas \cref{tab:main_results} retains their published scores.

\noindent\textbf{Checkpoints and depth selection.}
\method (SFT only) uses the final 16-frame SFT checkpoint, evaluated at fixed $r=2$, and \method (Ours) the final \alder checkpoint with adaptive stopping; the 32-frame rows evaluate the same checkpoints, gate, and threshold with only the frames increased from 16 to 32.
Checkpoints are not selected per benchmark, and fixed and adaptive results use identical weights and inputs.
For the random control in \cref{fig:slots_depth}b, let $p_{b,r}$ be the learned policy's depth frequency on benchmark $b$; its expected accuracy is $\sum_r p_{b,r}\operatorname{Acc}_{b,r}$, drawing depth independently of the question with the same depth budget.

\section{Complete Depth Analysis}
\label{sec:full_depth_analysis}

\begin{figure*}[t]
    \centering
    \includegraphics[width=\linewidth]{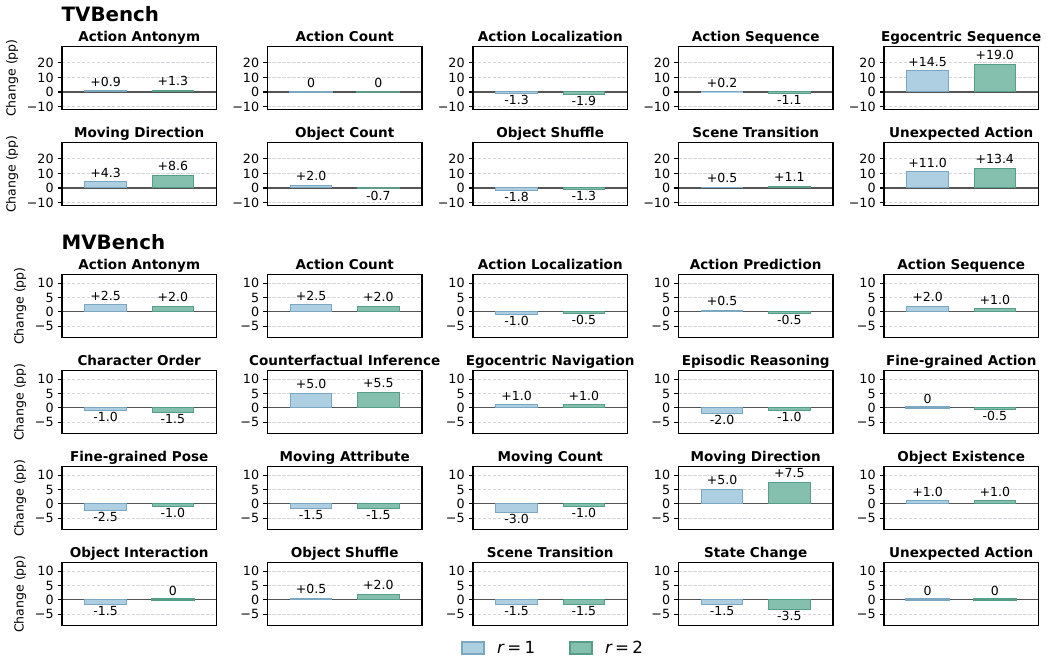}
    \caption{\textbf{Complete category-level depth analysis.}
All ten TVBench and twenty MVBench categories, ordered alphabetically.
Bars show accuracy changes (pp) at $r=1,2$ relative to $r=0$ for \method (SFT only) at fixed depths, before \alder, at 16 frames.
Scales are shared within each benchmark.}
    \label{fig:sft_depth_categories_full}

    \vspace{3mm}
    \captionof{table}{\textbf{Fixed-depth accuracy beyond the trained depth.} Accuracy (\%) at 16 frames from a separate evaluation of the same checkpoints; $^\ast$: beyond the trained depth.}
    \label{tab:depth_r3}
    \vspace{1.5mm}

    \footnotesize
    \setlength{\tabcolsep}{5pt}
\begin{tabular}{@{}lrrrrrrr@{}}
\toprule
\multicolumn{8}{c}{\method (SFT only)} \\
\cmidrule(lr){1-8}
$r$ & MVB & TC & TVB & VH & NQA & VME & Avg. \\
\midrule
0 & 68.2 & 74.8 & 54.8 & 40.0 & 80.4 & 58.7 & 62.8 \\
1 & 68.7 & 74.5 & 56.8 & 40.4 & 80.4 & 59.4 & 63.4 \\
2 & 69.0 & 74.5 & 57.3 & 40.6 & 80.4 & 59.6 & 63.5 \\
3$^\ast$ & 68.8 & 74.6 & 57.1 & 40.6 & 80.3 & 58.9 & 63.4 \\
\bottomrule
\end{tabular}
    \hspace{8mm}
\begin{tabular}{@{}lrrrrrrr@{}}
\toprule
\multicolumn{8}{c}{\method (Ours)} \\
\cmidrule(lr){1-8}
$r$ & MVB & TC & TVB & VH & NQA & VME & Avg. \\
\midrule
0 & 69.6 & 75.5 & 56.8 & 41.3 & 80.8 & 59.2 & 63.9 \\
1 & 69.9 & 75.6 & 57.2 & 41.5 & 80.8 & 59.7 & 64.1 \\
2 & 69.6 & 75.7 & 57.5 & 41.8 & 80.7 & 59.6 & 64.1 \\
3$^\ast$ & 69.5 & 75.5 & 57.2 & 42.0 & 80.7 & 59.3 & 64.0 \\
\bottomrule
\end{tabular}
\end{figure*}

\Cref{fig:sft_depth_categories_full} extends the depth analysis of \cref{sec:depth_diagnostic} to all ten TVBench and twenty MVBench categories. As in the main paper, the model is \method (SFT only), before \alder, evaluated at fixed depths.

\noindent\textbf{Depth beyond training.}
\method (SFT only) and \method (Ours) were trained with at most two loops. \Cref{tab:depth_r3} evaluates them separately at fixed depths $r=0$--$3$ with the same inputs; its $r\le2$ values agree with the main evaluation within 0.06 points on the six-benchmark mean.
At $r=3$ the mean falls below $r=2$ for both checkpoints, and from $r=2$ to $r=3$ more answers change from correct to incorrect than the reverse.

\end{document}